\documentclass[runningheads]{llncs}

\usepackage{eccv}

\usepackage{eccvabbrv}

\usepackage{graphicx}
\usepackage{booktabs}
\usepackage{multirow}
\usepackage{arydshln} 

\usepackage[accsupp]{axessibility}  

\usepackage{hyperref}

\usepackage{orcidlink}

\begin{document}

\title{OSR-ViT: A Simple and Modular Framework for Open-Set Object Detection and Discovery}

\titlerunning{OSR-ViT}

\author{Matthew Inkawhich\inst{1}
\and Nathan Inkawhich\inst{2}
\and Hao Yang\inst{1}
\and Jingyang Zhang\inst{1}
\and Randolph Linderman\inst{1}
\and Yiran Chen\inst{1}
}

\authorrunning{M.~Inkawhich et al.}

\institute{Duke University 
\and Air Force Research Laboratory
\\
}

\maketitle

\begin{abstract}

    An object detector's ability to detect and flag \textit{novel} objects during open-world deployments is critical for many real-world applications. Unfortunately, much of the work in open object detection today is disjointed and fails to adequately address applications that prioritize unknown object recall \textit{in addition to} known-class accuracy. To close this gap, we present a new task called Open-Set Object Detection and Discovery (OSODD) and as a solution propose the Open-Set Regions with ViT features (OSR-ViT) detection framework. OSR-ViT combines a class-agnostic proposal network with a powerful ViT-based classifier. Its modular design simplifies optimization and allows users to easily swap proposal solutions and feature extractors to best suit their application. Using our multifaceted evaluation protocol, we show that OSR-ViT obtains performance levels that far exceed state-of-the-art supervised methods. Our method also excels in low-data settings, outperforming supervised baselines using a fraction of the training data.
  
\end{abstract}

\section{Introduction}
\label{sec:introduction}
Traditional object detection models are designed, trained, and evaluated under closed-set conditions \cite{RCNN, FastRCNN, FasterRCNN, YOLO, SSD, DETR, DeformableDETR}, where all potential classes of interest are assumed to be exhaustively labeled in the training dataset. If such a model is deployed in an open-set environment \cite{OGOpenSet, OverlookedElephantOpenSet} where there exists unknown objects from outside the training class distribution, the model can either misclassify the object as a known class or miss the detection altogether -- leading to serious safety, equity and reliability concerns. This motivates the need for open-set object detection \cite{TowardsOpenWorldObjectDetection}, where unknown ``out-of-distribution'' (OOD) objects are explicitly handled in addition to the known ``in-distribution'' (ID) objects.

Although there have been many works that attempt to address open-set detection \cite{Miller1, Miller2, TowardsOpenWorldObjectDetection, VOS, SIREN, STUD}, we posit that the way they choose to handle unknown objects severely limits their practical usefulness. Namely, none of them consider \textbf{OOD object recall}. For example, seminal works by Miller et al. \cite{Miller1, Miller2} and Dhamija et al. \cite{OverlookedElephantOpenSet} define proper ``Open-Set Object Detection'' (OSOD) behavior as simply avoiding detecting any OOD objects as ID classes. More recent works by Du et al. \cite{VOS, SIREN, STUD} tackle ``Unknown-Aware Object Detection'' (UAOD), where the model is expected to accurately flag OOD objects that happen to be proposed to the detector's classifier head, but does not encourage OOD proposals. While these behaviors may be sufficient for some tasks, many applications require the explicit detection (i.e., discovery) of \textit{all} objects of interest, both ID and OOD.
For example, autonomous vehicles are often exposed to unforeseeable obstacles that demand detection for safe operation \cite{huang2019apolloscape, sun2020scalability}. Content moderation systems must also accurately identify evolving types of content while navigating the complexities of insufficient filtering \cite{moon2022difficulty}. Further, medical image processing models are relied upon to detect abnormalities \cite{deng2024medical}. In such cases, the consequences of poor OOD recall are severe, necessitating a new open-set task that prioritizes it.

\begin{figure}[t]
  \centering
   \includegraphics[width=0.8\linewidth]{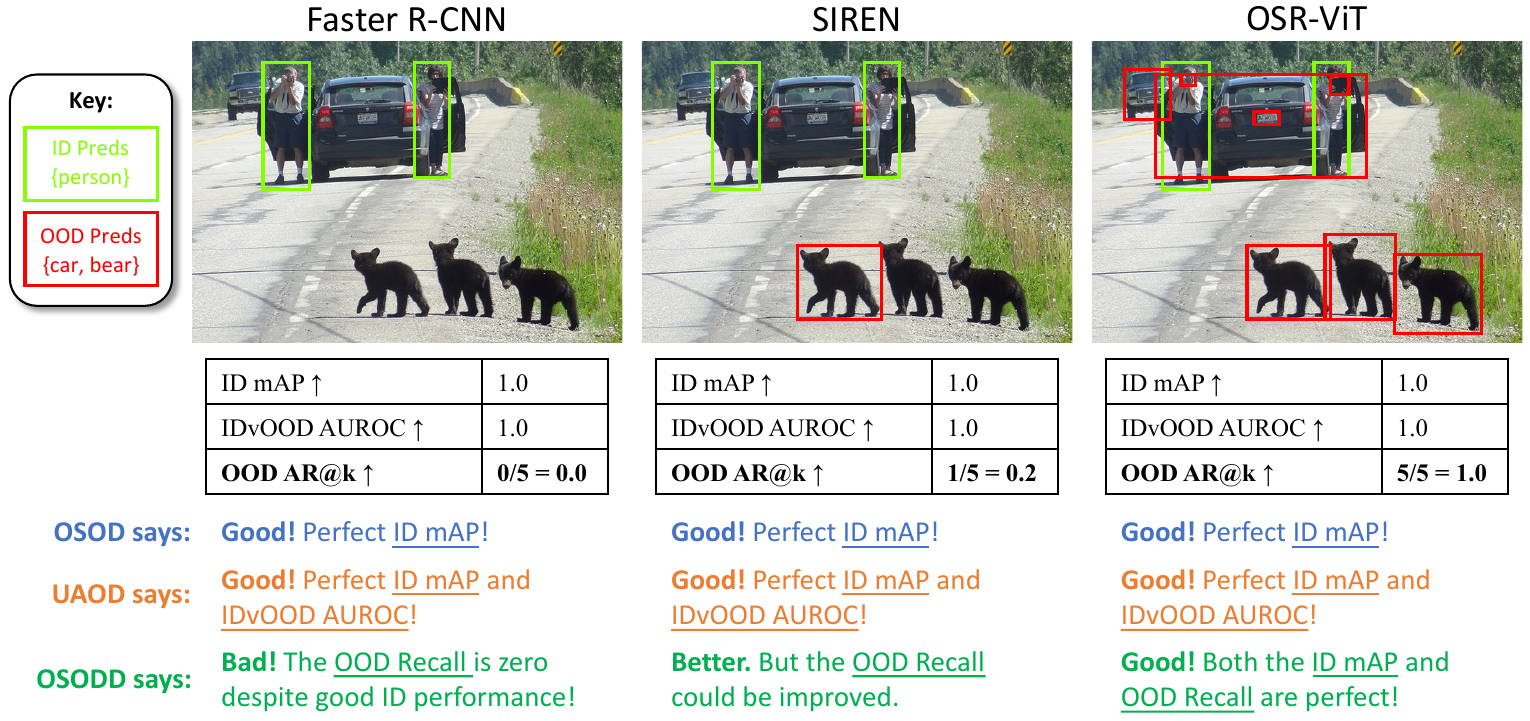}
   \vspace{-3mm}
   \caption{While other settings ignore OOD recall, the proposed OSODD task prioritizes it in addition to established metrics. In this example, a ``perfect'' model according to the OSOD or UAOD protocol may cause severe safety consequences.}
   \label{fig:evaluation_viz}
   \vspace{-5mm}
\end{figure}

In this work, we introduce \textbf{Open-Set Object Detection and Discovery (OSODD): a \underline{task} that explicitly prioritizes both ID-class accuracy \textit{and} OOD object recall}. OSODD more appropriately models many realistic applications like the ones mentioned above.
To measure performance on the OSODD task, we devise a new evaluation protocol that makes no simplifying assumptions about the test data and includes a novel threshold-independent Average Open Set Precision (AOSP) summary metric. We test models on three new benchmarks that are designed to simulate a broad spectrum of feasible settings, including low-data environments and multiple image domains. Not only does our OSODD evaluation protocol enable a more comprehensive analysis of model performance, but it also is the first that allows for a unified comparison of models from several subdivisions of open detection (e.g., OSOD \cite{OverlookedElephantOpenSet}, UAOD \cite{VOS}, Open-World Object Detection (OWOD) \cite{TowardsOpenWorldObjectDetection}). Such a comparison highlights how poorly those solutions perform in OSODD (see Fig. \ref{fig:evaluation_viz}).

To address the OSODD task, we create a new \textbf{highly-modular detection framework, called Open Set Regions with ViT features (OSR-ViT)}. This framework is comprised of a dedicated class-agnostic proposal network combined with a classifier module that leverages powerful off-the-shelf ViT-based foundational models. OSR-ViT's bipartite architecture does not require end-to-end training, so users can easily replace either of the components with future or custom models. In this paper's instantiation of the framework we use the state-of-the-art Tunable Hybrid Proposal Network (THPN) \cite{THPN} and DINOv2 \cite{DINOv2} foundational model. 
We find that our simple, modular, and user-friendly OSR-ViT framework far exceeds the performance of all fully supervised open-set-specific baselines. Our framework has a particular advantage in low-data settings, where even our most lightweight configuration trained on 25\% of the PASCAL VOC \cite{PASCALVOC} training data outperforms all other baselines trained on 100\% of the data. 

Overall, our contributions are as follows:
\begin{itemize}
\vspace{-0.3em}
    \item We create a new \textbf{joint open-set object detection and discovery task} that prioritizes both ID and OOD object detection and is more closely aligned with realistic open-set detection applications. To measure performance we develop a new comprehensive evaluation protocol and AOSP summary metric that allows for a unified comparison of previously uncompared works.\\ 
\vspace{-0.6em}
    \item  We propose a novel OSR-ViT framework for tackling the OSODD task. OSR-ViT's modularity allows for the immediate use of the latest foundational models being developed, future-proofing our design. \\ 
\vspace{-0.6em}
    \item  We show that OSR-ViT vastly outperforms fully-supervised alternatives with minimal configuration and no finicky end-to-end training. We also demonstrate its effectiveness with sparse-data and in the remote-sensing domain.
\end{itemize}
\vspace{-1em}

\section{Limitations of Existing Work}
\label{sec:related_work}

A major limitation of existing ``open'' object detection literature is the incongruity of task goals. The evaluation protocols used in different works vary significantly, making it very challenging to directly compare methods. Here, we detail the existing subdivisions of open detection along with their limitations.

\textbf{Class-Agnostic Object Proposal.} 
Object discovery models separate objects from background without supervision \cite{OGObjectDiscovery}. Early works in this area identify salient regions with respect to image transformations \cite{OGObjectDiscovery, ObjectDiscoveryInTheWild} or noise \cite{ObjectDiscoveryRGBD}. More recent works leverage convolutional features instead of the images directly \cite{DissimilarityCoeffBasedObjDiscovery, ObjectCentricLearningSlotAttention}.
Class-agnostic object proposal networks seek to maximize ID and OOD object recall (without further classification) \cite{KevinsPaper, OLN, LearningToDetectEveryThing, THPN}. Kim et al. \cite{OLN} showed that standard object proposal networks such as Region Proposal Network (RPN) \cite{FasterRCNN} and its variants \cite{CascadeRPN, RegionProposalGuidedAnchoring} overfit to the ID categories because of its discriminative learning approach. They instead propose an Object Localization Network (OLN) which replaces the classification heads of a class-agnostic Faster R-CNN \cite{FasterRCNN} with localization-quality prediction heads, yielding a model that more readily generalizes to OOD objects. Konan et al. \cite{KevinsPaper} and Saito et al. \cite{LearningToDetectEveryThing} use unknown object masking and a background erasing augmentation, respectively, to further reduce ID-bias. While class-agnostic detection is useful, downstream class separation is often necessary for practical tasks.

\textbf{Open-Set \& Unknown-Aware Object Detection.} 
An OSOD detector should ignore OOD objects and not let the presence of OOD or ``wilderness'' data effect ID accuracy \cite{OverlookedElephantOpenSet, Miller1, Miller2, Miller3, ExpandingLowDensityLatentRegionsOSOD}. In other words, the goal is to simply avoid mistaking OOD objects as ID classes.
Miller et al. \cite{Miller1} first introduce the notion of open-set object detection and use dropout sampling \cite{DropoutSampling} to improve label uncertainty.
Dhamija et al. \cite{OverlookedElephantOpenSet} show that closed-set detectors tend to misclassify OOD objects as ID classes. 
Recently, Han et al. \cite{ExpandingLowDensityLatentRegionsOSOD} use a contrastive feature learner to identify OOD objects from their latent representations. The limitation of OSOD is that the recall of OOD objects is irrelevant, which render these methods unfit for many real applications. 
An UAOD model should maximize ID performance and precisely flag any OOD objects that happen to be proposed to the classifier head \cite{VOS, SIREN, STUD}. 
Du et al. \cite{VOS} generate near-OOD virtual outliers to learn more compact ID clusters to ease the separation of ID and OOD objects. SIREN \cite{SIREN} maps ID-class representations onto a von Mises-Fisher (vMF) distribution to provide a powerful distance-based OOD algorithm for detectors. Finally, STUD \cite{STUD} distills unknown objects from video data to improve OOD detection in object detection models.
A major limitation of this subdivision is that most works \cite{VOS, SIREN, STUD} make several unrealistic and invalid assumptions to evaluate performance in the detection task. For example, they require mutually exclusive ID and OOD validation sets, and incorrectly assume that all detections with confidence over a certain threshold are valid ID and OOD predictions, respectively. 

\textbf{Open-World \& Open-Vocabulary Object Detection.}
An OWOD model's goal is to maximize ID performance and incrementally learn new classes by forwarding it's unknown predictions to a human annotator \cite{TowardsOpenWorldObjectDetection, OWDETR, RevisitingOWOD, UCOWOD, ObjectsInSemanticTopology, OWODDiscriminativeClassPrototype, TwoBranchObjCentricOWOD, PROB}. Joseph et al.'s ORE model \cite{TowardsOpenWorldObjectDetection} uses a conventional RPN with a contrastive clustering regularization to create a baseline. Gupta et al. \cite{OWDETR} introduce a DETR-based \cite{DETR, DeformableDETR} OW-DETR model that boosts performance via attention-driven pseudo-labeling. Wu et al. \cite{TwoBranchObjCentricOWOD} propose a two-branch objectness-centric model that leverages the benefits of OLN's localization-quality prediction head to improve object recall. Finally, Zohar et al.'s PROB model \cite{PROB} specifically addresses unknown object recall with an additional probabilistic objectness head.
While the incremental learning aspect of the OWOD task is interesting, several outside factors (e.g., threshold choice, semantic drift between tasks, training data quality) contribute heavily to perceived performance, making it difficult to judge a model's true usability. Also, while some work does enhance OWOD performance via increased unknown recall \cite{TwoBranchObjCentricOWOD, PROB}, their OOD recall performance still remains modest.
Open-Vocabulary Object Detection (OVOD) models use natural language models to enable the detector to directly generalize beyond the ID classes using text prompting \cite{OVODusingCaptions, OVDETR, ObjectCentricOVD, CORA, ScalingOVOD}. While these approaches are powerful under certain circumstances (i.e., where object classes of interest are well-represented in language datasets), their practical usefulness is limited in many domains (i.e., fine-grained ship detection). For fair comparison, we do not consider OVOD baselines as they involve a very different multi-modal approach.

\section{Open Set Object Detection and Discovery}
\label{sec:osodd}

\begin{figure}[t]
  \centering
   \includegraphics[width=0.8\linewidth]{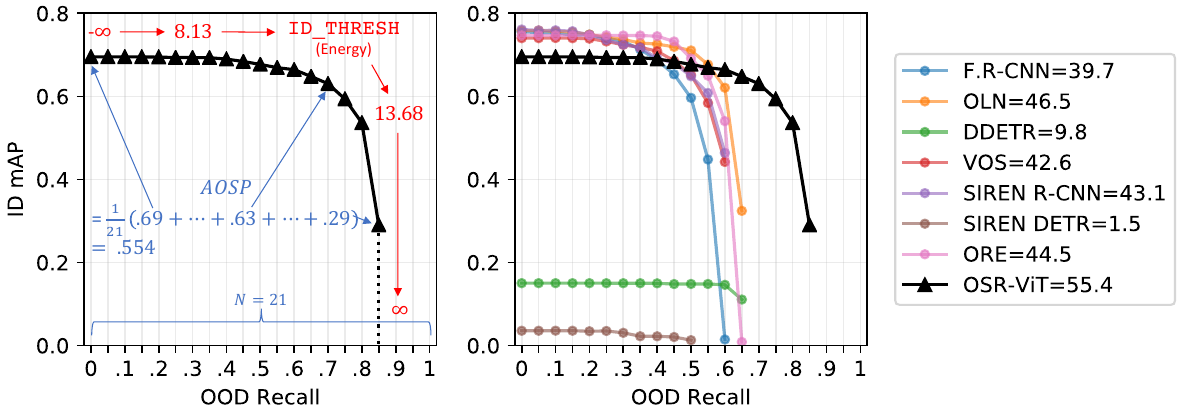}
   \vspace{-3mm}
   \caption{Our threshold-agnostic Average Open Set Precision (AOSP) performance metric provides a holistic view of the ID-OOD performance trade-off.}
   \label{fig:aosp_metric}
   \vspace{-3mm}
\end{figure}

In this section we describe in detail the OSODD task.
In Sec. \ref{subsec:osodd_problem_formulation} we formalize the problem with notation and in Sec. \ref{subsec:osodd_evaluation_protocol} we detail our novel performance metrics and evaluation protocol.

\subsection{Problem Formulation}
\label{subsec:osodd_problem_formulation}
As with any supervised detection task, we assume access to a training dataset that contains labels for a set of object classes of interest. We refer to this set of classes as the \textit{known} set $\mathcal{K} = \{1, 2, \dots, C\} \subset \mathbb{N}^+$. In OSODD, we also formally acknowledge the existence of instances of \textit{unknown} object classes $\mathcal{U} = \{C+1, \dots \} \subset \mathbb{N}^+$ that coexist with the known instances in both the training and deployment data. The goal is to train a model $\mathcal{M}$ parameterized by $\theta$ to detect and localize \textit{all} object instances of interest in a test set (i.e., all instances in the set $\mathcal{K} \cup \mathcal{U}$). For a given test image $X$, the model's function is $\mathcal{M}(X; \theta) = \{[x,y,w,h,c,s]_{i=1\dots N}\}$, where $x$, $y$, $w$, and $h$ denote the center coordinates, width, and height of the bounding box, respectively. The predicted class $c \in \mathcal{K} \cup \{-1, 0\}$ describes the class category that the $i$th prediction belongs to. Here, $c=0$ denotes an \textit{unknown} object of interest and $c=-1$ represents \textit{background} (i.e., no object). Finally, each prediction has a score $s \in [0,1]$ which represents the model's confidence that box $i$ contains an object of class $c$.

\subsection{Evaluation Protocol}
\label{subsec:osodd_evaluation_protocol}
A key contribution of our work is the novel evaluation procedure we develop for the OSODD task. 
Our evaluation uses four types of metrics to comprehensively evaluate models without any unrealistic assumptions or specific thresholding:
\begin{itemize}
    \item \textbf{Closed-Set ID mean Average Precision (ID-mAP)}: Measures the maximum potential ID-mAP by assuming all detections are knowns.
    \item \textbf{Class-Agnostic Average Recall (CA-AR)}: Isolates the performance of the proposal network by measuring the AR@100 assuming a single foreground (FG) class.
    \item \textbf{Area Under the Receiver Operating Characteristic (AUROC)}: Measures the classifier's separability across all possible thresholds. Recent UAOD works \cite{VOS, SIREN, STUD} only measure ID vs. OOD AUROC, since they assume that an input is always either ID or OOD (binary). However, such an assumption is inadequate for the OSODD task where we face a ternary decision: A proposal can either be an ID object, an OOD object, or background (BG). Thus, we also compute AUROC for the following separation axes: ID vs. Non-ID, OOD vs. BG, and FG vs. BG. 
    \item \textbf{Average Open-Set Precision (AOSP)}: Our new AOSP metric provides a threshold-independent summary of a model's tradeoff between ID-mAP and OOD Recall. This metric is described in detail below.
\end{itemize}

\textbf{Computing AUROC.}
Unlike existing works \cite{VOS, SIREN} that use AUROC for open-set detection, we do \textit{NOT} require that ID and OOD data are in mutually exclusive sets, and we do \textit{NOT} assume that all high-confidence predictions are valid object regions. Instead, we take a more scrupulous approach and partition all proposed regions in the mixed test set (i.e., the images contain both ID and OOD objects) into their corresponding ID/OOD/BG bin based on their IoU overlap with the ground-truth annotations. Note that during evaluation, we always pretend that some subset of classes are OOD, so we have ground truth matches for OOD objects too.
Once the predictions are partitioned, we compute our AUROC scores. ID vs. OOD and ID vs. Non-ID AUROC are computed using the proposal's ID score, which should be high for ID objects and low for OOD objects (e.g., energy \cite{EnergyOOD}, Mahalanobis distance \cite{Mahalanobis}, etc.). BG vs. OOD and FG vs. BG AUROC are computed using the objectness score, which represents the likelihood that a region contains a foreground object (either ID or OOD).

\textbf{The AOSP metric.}
While it is tempting to try to use mAP to measure OOD performance, this is invalid because computing \textit{precision} requires that \textit{all} OOD objects are labeled. Due to limitations of current datasets, we do not have exhaustive annotations for every single object. Thus, the accepted standard for measuring OOD performance is \textit{recall} given a fixed number $k$ of detections per image. However, we argue that the true performance of an OSODD model cannot be fully understood from a single recall measure, as it only captures performance at a one operating point. This point is determined by a model's \texttt{ID\_THRESH}, the threshold which determines the minimum ID score for a prediction to be deemed an ID object. We argue that the best way to evaluate a model is to use a threshold-independent metric that summarizes the tradeoff between ID and OOD performance, as different applications require different thresholds. 

To this end, we propose \textbf{Average Open-Set Precision (AOSP)}. 
AOSP summarizes the tradeoff between ID-mAP (@IoU=0.5) and OOD recall (@$k$=100 detections per image), and provides us with a single scalar metric to compare methods on the OSODD task. Figure \ref{fig:aosp_metric} shows a visualization of the AOSP computation. 
We specifically find the minimum \texttt{ID\_THRESH} to achieve 21 discrete target OOD recall points in $\{0\!:\!.05\!:\!1\}$. At each of these, we set $c$=0 (\textit{unknown}) for all detections with ID score $<$ \texttt{ID\_THRESH} and compute ID-mAP on the updated set. AOSP is the average of ID-mAP over these OOD recall points:
\begin{equation} \label{eq:aosp}
\mathrm{AOSP} := \frac{1}{21} \sum_{r \in \{0:.05:1\}} \mathrm{ID\text{-}mAP}_{@\mathrm{OOD\ Recall\ =\ }r}
\end{equation}
Note that at $\texttt{ID\_THRESH}$=$-\infty$ every detection is deemed ID (max ID-mAP), and at $\texttt{ID\_THRESH}$=$\infty$ every detection is deemed OOD (max OOD recall). At OOD recall points beyond the detector's maximum capability (e.g., $r$=\{0.9,0.95,1\} in Fig. \ref{fig:aosp_metric}), we consider ID-mAP=0.

\section{OSR-ViT Modular Detection Framework}
\label{sec:methodology}

\begin{figure}[t]
  \centering
   \includegraphics[width=1.0\linewidth]{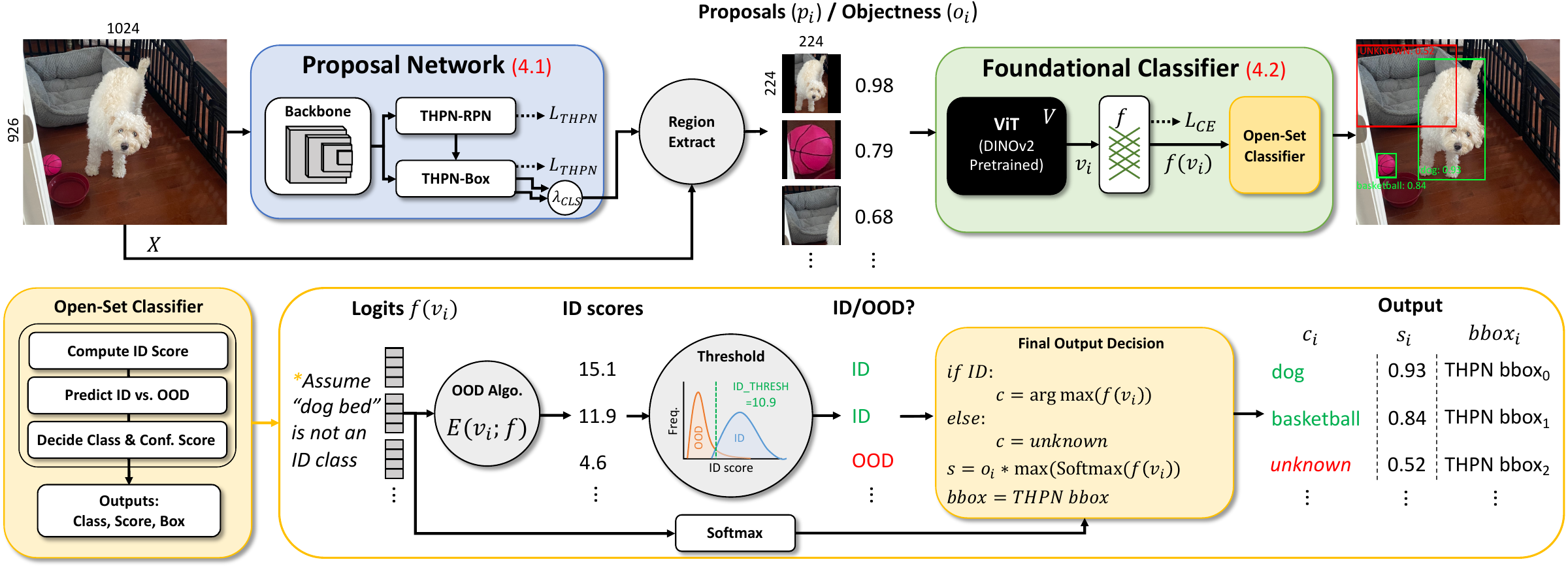}
   \vspace{-6mm}
   \caption{Our OSR-ViT framework consists of two independently-trained models working in conjunction: (1) a class-agnostic Proposal Network, and (2) a ViT-powered Foundational Classifier. This allows for seamless integration of new or future models.}
   \label{fig:architecture}
   \vspace{-2mm}
\end{figure}

An effective OSODD model must excel at two key subtasks: (1) localizing all objects in an image, and (2) accurate discernment between ID and OOD classes. Thus, our proposed solution is a modular bipartite framework that combines an arbitrary strong proposal network with a classifier module that leverages an arbitrary foundational Vision Transformer (ViT) \cite{ViT} (see Fig. \ref{fig:architecture}).
\textbf{An important reason for this design choice is that in today's fast-paced ML climate, modularity is critical for future-proofing}. New state-of-the-art models are being released almost daily, necessitating frameworks that allow for seamless transitioning between solutions. 
The task-agnostic nature of these foundational models is also critical to being adaptable to dynamic environments and tasks. 
This is opposed to developing highly task-specific solutions that require extra hyperparameters, regularization terms, and underlying assumptions.
We call our solution Open-Set Regions with ViT features (OSR-ViT), taking inspiration from the seminal ``Regions with CNN features'' (R-CNN) model family \cite{RCNN, FastRCNN, FasterRCNN, MaskRCNN}. The remainder of this section details the Proposal Network (Sec. \ref{subsec:proposal_network}), the Foundational Classifier (Sec. \ref{subsec:foundational_classifier}), and model training (Sec. \ref{subsec:training}).

\subsection{Proposal Network}
\label{subsec:proposal_network}
The upper bound of overall OSODD performance is directly predicated on the model's ability to discern foreground objects vs. background, as even a detector with a perfect classifier is useless if true positive regions are never proposed in the first place. 
One major pitfall of open object proposal is overfitting to ID classes. 
Basic supervised proposal networks like RPN \cite{FasterRCNN} inherently overfit due to their discriminative objective \cite{OLN}. Several recent works have tried to combat this issue \cite{OLN, KevinsPaper}, however it has been shown that incorporating such dedicated proposal networks directly into end-to-end open-set/world detectors yields worse overall performance \cite{TwoBranchObjCentricOWOD, PROB}.
The other major pitfall is a lack of adaptability. A practically useful OSODD proposal network should be able to be adapted to different application requirements \cite{THPN}. For example, a security system should prioritize the detection of a couple of key ID classes (e.g., person, car) while ignoring unrelated OOD objects. However, a household robot should be much more generalizable to rare and unexpected object classes. 

In our instantiation of OSR-ViT, we use a Tunable Hybrid Proposal Network (THPN) \cite{THPN}. THPN is a state-of-the-art CNN-based proposal network that learns a hybrid objectness representation via dual prediction heads. 
Critically, THPN provides a single hyperparameter $\lambda_{CLS} \in [0,1]$ which balances both the loss contribution and final confidence score from each prediction head. 
The larger $\lambda_{CLS}$ is set, the more ID-biased the resulting model is, meaning the more propensity the model has for detecting ID objects at the cost of some OOD objects.
THPN also leverages a self-training optimization procedure \cite{Lee_pseudolabel} that significantly enhances data efficiency, allowing for impressive performance in low-data or semi-supervised settings.
We emphasize that OSR-ViT users can seamlessly plug-and-play with any proposal model of their choosing. For example, if an organization has developed an exquisite proposal network for a specific remote-sensing task, that network can be leveraged here.

\subsection{Foundational Classifier}
\label{subsec:foundational_classifier}
The recent emergence of large-scale foundational models has begun to revolutionize the pipeline of training and deploying vision AI. Open-source models such as CLIP \cite{CLIP} and DINOv2 \cite{DINOv2} are trained on hundreds of millions of images for tens of thousands of GPU hours. They provides users with off-the-shelf task-agnostic models that can be made task-specific with a minimal fine-tuning stage, and outperform supervised specialist models. This strong performance is due to the highly expressive representations that are encoded by the Vision Transformer (ViT) architecture. However, we argue that the true power of these foundational models extends far beyond closed-set recognition.
\textbf{Our hypothesis is that the highly descriptive ViT representations of the object proposals will enable effective ID \textit{and} OOD separation}.
In this work, we use a DINOv2 \cite{DINOv2} model as the feature extractor of the foundational classifier in the OSR-ViT. 
DINOv2 is trained on the extensive LVD-142M dataset \cite{DINOv2}, meaning it is fully capable of well-representing a wide variety of image domains and object types. Again, we encourage users to plug-and-play beyond DINOv2 with whatever new or custom foundational model they see fit.

As shown in Fig. \ref{fig:architecture}, the input image is first processed by the proposal network $\mathrm{PropNet}(X): \mathbb{R}^D \rightarrow \{(p_i^{bbox}, o_i)\}_{i=1}^{100}$, which maps a $D$-dimensional input image $X$ to $N$=100 pairs of object proposal boxes $p_i^{bbox}$ and their corresponding predicted objectness $o_i$. The pixel region of each proposal is then cropped from the image and resized to the 224x224 resolution that the DINOv2 model can ingest. We call these resulting resized proposal ``images'' $p_i$.
The proposal images are then forwarded through the ViT feature extractor $V(p) : \mathbb{R}^{224\cdot224} \rightarrow \mathbb{R}^{d}$, where $d$ is the dimensionality of the ViT's feature space. We refer to the ViT representation of proposal $p_i$ as $v_i$.
We use a simple 2-layer fully connected (non-linear) module $f(v) : \mathbb{R}^d \rightarrow \mathbb{R}^C$ on top of the ViT feature extractor to enable $C$-way classification. The output logits of each proposal $f(v_i)$ are then forwarded to the Open-Set Classifier which makes the final output decision. 

Reaching a final detection involves making two sequential predictions. First, we must predict if a proposal is ID or OOD. 
We opt for a post-hoc Energy-based OOD detection algorithm \cite{EnergyOOD} that uses a proposal's free energy as its ID score:
\begin{equation} \label{eq:energy}
E(v_i; f) = -T \cdot \log \sum_{j}^{C}e^{f_j(v_i)/T}
\end{equation}
where $T$ is a temperature parameter. Note that for a given proposal, the larger this energy score is, the more likely it is to be an ID class object. If $-E(v_i; f) > \texttt{ID\_THRESH}$ we call the $i$th proposal an ID object, else we call it \textit{unknown}. For deployment, one would use a validation set to choose a reasonable \texttt{ID\_THRESH}.

The second decision that must be made by the Open-Set Classifier is the final output class $c_i$ and confidence score $s_i$. Fig. \ref{fig:architecture} shows this decision in the ``Final Output Decision'' box. If $p_i$ is deemed OOD, the assigned class label is $c_i$~= \textit{unknown}, but if it is deemed ID then $c_i=\arg\max(f(v_i))$.
Regardless of class label, the confidence score is the product of the predicted objectness from the proposal network $o_i$ and the max Softmax confidence over the ID classes:
\begin{equation} \label{eq:score}
s_i=o_i \cdot \max(\text{Softmax}(f(v_i))). 
\end{equation}
 Note that many existing works \cite{VOS, SIREN, STUD} simply use the maximum Softmax score for OOD predictions. Although this may be valid for the binary open-set \textit{classification} task (ID vs. non-ID), it is not appropriate for the ternary open-set \textit{detection} task (ID vs. OOD vs. BG). In other words, \textbf{just because a proposal does not significantly excite any one ID output node does NOT necessarily mean that it does not have strong general object features}. For this reason, our score measure $s_i$ directly incorporates the objectness score from the class agnostic proposal network, meaning the resulting scores for both ID and OOD predictions will be more appropriately calibrated.
 Finally, we reuse the bounding boxes predicted by the proposal network as the final box predictions.

\subsection{Training}
\label{subsec:training}
Much of OSR-ViT's user-friendliness is due to its disentangled training of the proposal network and the foundational classifier. This allows users to easily incorporate new custom or off-the-shelf models for either role. In this work, we optimize the THPN following the procedure outlined in the paper \cite{THPN}. We adapt the foundational classifier separately, and in two stages. In the first stage, we freeze the DINOv2-pretrained ViT and update the fully connected classifier module $f$ using cross-entropy loss for 50 epochs. To improve model flexibility while maintaining the expressiveness of the ViT's pre-trained representations, we then perform a short 5-epoch fine-tuning stage in which we train the ViT and the classifier module together with a much smaller learning rate. Specific implementation details and hyperparameters can be found in Appendix A. 

\section{Experiments}
\label{sec:experiments}
\vspace{-1mm}
To evaluate models on the OSODD task, we create three separate benchmarks which offer far more diversity than contemporary literature \cite{Miller1, OverlookedElephantOpenSet, VOS, SIREN, ExpandingLowDensityLatentRegionsOSOD, Miller3}. Sec. \ref{subsec:natural_imagery_benchmarks} contains our Natural Imagery Benchmark, Sec. \ref{subsec:limited_data_benchmark} contains our Limited Data Benchmark, and Sec. \ref{subsec:ships_benchmark} covers model performance on the Ships Benchmark. Finally, in Sec. \ref{subsec:osrvit_performance_analysis} we perform additional analysis on our OSR-ViT method. 

\vspace{-2mm}
\subsection{Natural Imagery Benchmark}
\label{subsec:natural_imagery_benchmarks}
\begin{table}[t]
\caption{Results on the Natural Imagery Benchmark tasks.}
\vspace{-3mm}
\centering
\resizebox{0.95\linewidth}{!}{
\begin{tabular}{l|l|l|lccccccc}
\toprule
                                                              &                                                      &                                  &              &               &               &               & IDvOOD         & IDvNONID       & OODvBG         & FGvBG          \\
Data                                                          & Training                                             & Model                            & OOD Algo.    & AOSP        & ID-mAP        & CA-AR         & AUROC          & AUROC          & AUROC          & AUROC          \\ \hline\hline &&&\tabularnewline[-2.2ex]
\multirow{15}{*}{\shortstack[l]{VOC $\rightarrow$\\ COCO}}    & \multirow{6}{*}{\shortstack[l]{Plain\\ Supervised}} & \multirow{2}{*}{Faster R-CNN}    
& Energy       & 17.8          & 31.1          & 37.2          & 73.41          & 64.00          & 59.77          & 65.64          \\
                                                              &                                                      &                                  & Mahalanobis  & 18.0          & 31.1          & 37.2          & 56.27          & 68.32          & 59.77          & 65.64          \\ \cline{3-11} &&&\tabularnewline[-2.2ex]
                                                              &                                                      & \multirow{2}{*}{OLN}             
& Energy       & 18.8          & 30.0          & 38.5          & 72.42          & 64.94          & 59.44          & 66.29          \\
                                                              &                                                      &                                  & Mahalanobis  & 18.4          & 30.0          & 38.5          & 51.66          & 65.81          & 59.44          & 66.29          \\ \cline{3-11} &&&\tabularnewline[-2.2ex]
                                                              &                                                      & \multirow{2}{*}{Deformable DETR} 
& Energy       & 10.1          & \textbf{34.6} & 33.3          & 58.77          & 69.05          & 58.58          & 57.62          \\
                                                              &                                                      &                                  & Mahalanobis  & 9.8           & \textbf{34.6} & 33.3          & 55.25          & 63.35          & 58.58          & 57.62          \\ \cline{2-11} &&&\tabularnewline[-2.2ex]
                                                              & VOS                                                  & Faster R-CNN                     & Energy       & 18.6          & 31.5          & 36.3          & 78.68          & 73.55          & 61.44          & 73.68          \\ \cline{2-11} &&&\tabularnewline[-2.2ex]
                                                              & \multirow{2}{*}{SIREN}                               & Faster R-CNN                     & SIREN-KNN    & 17.3          & 31.3          & 36.7          & 82.74          & 77.15          & 58.91          & 64.23          \\
                                                              &                                                      & Deformable DETR                  & SIREN-KNN    & 12.0          & 33.6          & 33.1          & 75.87          & 82.70          & 57.98          & 57.74          \\ \cline{2-11} &&&\tabularnewline[-2.2ex]
                                                              & ORE                                                  & Faster R-CNN                     & Energy       & 18.3          & 28.0          & 35.4          & 75.13          & 74.90          & 53.91          & 63.01          \\ \cline{2-11} &&&\tabularnewline[-2.2ex]
                                                              & OW-DETR                                              & Deformable DETR                  & Direct Pred. & 10.7          & 30.2          & 30.9          & -              & -              & -              & -              \\ \cline{2-11} &&&\tabularnewline[-2.2ex]
                                                              & PROB                                                 & Deformable DETR                  & Direct Pred. & 12.6          & 32.5          & 31.7          & -              & -              & -              & -              \\ \cline{2-11} &&&\tabularnewline[-2.2ex]
                                                              & \multirow{3}{*}{OSR-ViT}                                & THPN+DINOv2-S                    & Energy       & 23.6          & 30.2          & \textbf{43.2}          & 84.79          & 85.08          & 63.26          & 80.69          \\
                                                              &                                                      & THPN+DINOv2-B                    & Energy       & 25.0          & 31.4          & \textbf{43.2} & 86.49          & \textbf{86.28} & 63.42          & 81.86          \\
                                                              &                                                      & THPN+DINOv2-L                    & Energy       & \textbf{25.1} & 31.5 & \textbf{43.2} & \textbf{87.57} & 85.52          & \textbf{64.87} & \textbf{82.31} \\ \hline\hline &&&\tabularnewline[-2.2ex]
\multirow{13}{*}{\shortstack[l]{COCO $\rightarrow$\\ Obj365}} & \multirow{6}{*}{\shortstack[l]{Plain\\ Supervised}} & \multirow{2}{*}{Faster R-CNN}    
& Energy       & 17.6          & 24.5          & 44.1          & 61.84          & 65.10          & 63.62          & 66.99          \\
                                                              &                                                      &                                  & Mahalanobis  & 14.6          & 24.5          & 44.1          & 53.71          & 56.97          & 63.62          & 66.99          \\ \cline{3-11} &&&\tabularnewline[-2.2ex]
                                                              &                                                      & \multirow{2}{*}{OLN}             
& Energy       & 17.5          & 23.0          & 44.9          & 62.66          & 65.06          & 63.25          & 66.32          \\
                                                              &                                                      &                                  & Mahalanobis  & 13.6          & 23.0          & 44.9          & 52.39          & 56.60          & 63.25          & 66.32          \\ \cline{3-11} &&&\tabularnewline[-2.2ex]
                                                              &                                                      & \multirow{2}{*}{Deformable DETR} 
& Energy       & 17.3          & \textbf{29.0} & 43.9          & 55.57          & 60.50          & 58.04          & 61.34          \\
                                                              &                                                      &                                  & Mahalanobis  & 13.2          & \textbf{29.0} & 43.9          & 48.11          & 46.98          & 58.04          & 61.34          \\ \cline{2-11} &&&\tabularnewline[-2.2ex]
                                                              & VOS                                                  & Faster R-CNN                     & Energy       & 17.8          & 24.4          & 43.6          & 65.20          & 68.16          & 63.25          & 67.34          \\ \cline{2-11} &&&\tabularnewline[-2.2ex]
                                                              & \multirow{2}{*}{SIREN}                               & Faster R-CNN                     & SIREN-KNN    & 17.0          & 24.4          & 43.4          & 68.34          & 68.68          & 62.91          & 66.99          \\
                                                              &                                                      & Deformable DETR                  & SIREN-KNN    & 8.2           & 28.8          & 43.4          & 71.45          & 73.75          & 58.43          & 60.75          \\ \cline{2-11} &&&\tabularnewline[-2.2ex]
                                                              & ORE                                                  & Faster R-CNN                     & Energy       & 16.9          & 22.7          & 42.4          & 62.35          & 66.17          & 60.09          & 64.07          \\ \cline{2-11} &&&\tabularnewline[-2.2ex]
                                                              & \multirow{3}{*}{OSR-ViT}                                & THPN+DINOv2-S                    & Energy       & 18.7          & 23.9          & \textbf{49.7}          & 67.01          & 73.57          & 68.55          & 73.89          \\
                                                              &                                                      & THPN+DINOv2-B                    & Energy       & 19.7          & 25.1          & \textbf{49.7} & 70.72          & 75.81          & 67.16          & 73.70          \\
                                                              &                                                      & THPN+DINOv2-L                    & Energy       & \textbf{20.2} & 25.7          & \textbf{49.7} & \textbf{71.60} & \textbf{76.67} & \textbf{67.33} & \textbf{74.04} \\
\bottomrule
\end{tabular}

}
\label{tab:main_voccoco_cocoobj365}
\end{table}
This benchmark considers two cross-dataset transfer tasks between common natural imagery datasets. The first is to train on the 20-class PASCAL VOC \cite{PASCALVOC} training dataset and test on the 80-class COCO \cite{COCO} validation set. In this case, the OOD classes are the non-VOC classes of COCO. The second is to train on the COCO training set and test on 40,000 images from the 365-class Objects365 \cite{Objects365} dataset. Here, the OOD classes are the non-COCO classes of Objects365. Since the Objects365 label space is more granular we consider all synsets or hyponyms of the COCO classes as ID. 
Table \ref{tab:main_voccoco_cocoobj365} contains the results for this benchmark. Note that the ``-S'', ``-B'', and ``-L'' specifiers on the DINOv2 models indicate the size of the ViT.
Our OSR-ViT method outperforms all baselines on all OOD-related metrics on both tasks. In general, OSR-ViT's margin of improvement over the baselines is greater on VOC$\rightarrow$COCO compared to COCO$\rightarrow$Objects365. This is because the stronger supervised baselines (e.g., DETR-based models) can learn better representations of the ID classes in tasks with more training data. 

OSR-ViT significantly outperforms all baselines in terms of CA-AR, showcasing the utility of a non-ID-biased proposal network like THPN.
The relatively mediocre AOSP and CA-AR scores from the major OWOD methods (ORE \cite{TowardsOpenWorldObjectDetection}, OW-DETR \cite{OWDETR}, and PROB \cite{PROB}) shows that the incremental learning aspect of the OWOD task does indeed distract from the relatively poor OOD recall, justifying the need for our OSODD task.
Finally, OSR-ViT excels in terms of classifier separability (i.e., AUROC metrics). 
The strong ID score-based separation (ID vs. OOD, ID vs. Non-ID) demonstrates that ViT's strong nuanced representations allow superior OOD detectability, even compared to strong regularized UAOD baselines such as VOS \cite{VOS} and SIREN \cite{SIREN} that are specifically designed for this capability. The objectness-based separation (OOD vs. BG, FG vs. BG) is also much better than the baselines, with the FG vs. BG AUROC being 16.02\% higher than the best baseline (OLN). 

As expected, the size of the DINOv2 ViT does positively correlate with performance, but even DINOv2-S can provide state-of-the-art performance on both tasks in terms of AOSP. On the moderately-scaled VOC$\rightarrow$COCO task, the smallest DINOv2-S is still sufficient to outperform the UAOD methods in terms of classifier separability, but on the larger COCO$\rightarrow$Objects365 task the larger DINOv2-L is required to beat SIREN-DETR \cite{SIREN} in terms of ID vs. OOD AUROC. 
One limitation of our particular OSR-ViT configuration is that it trades off far superior OOD recall for slightly worse closed-set ID-mAP. Our analysis shows that this is not due to Foundational Classifier error, but rather to the ID/OOD tradeoff made by the THPN proposal network. In other words, we configure the THPN in these experiments with $\lambda_{CLS}$=.10, which yields a model with more OOD-bias. In Appendix D we explore the impact of $\lambda_{CLS}$ and show that the ID-mAP discrepancy can indeed be minimized.

\vspace{-2mm}
\subsection{Limited Data Benchmark}
\label{subsec:limited_data_benchmark}
\vspace{-1mm}

\begin{figure}[t]
  \centering
   \includegraphics[width=0.95\linewidth]{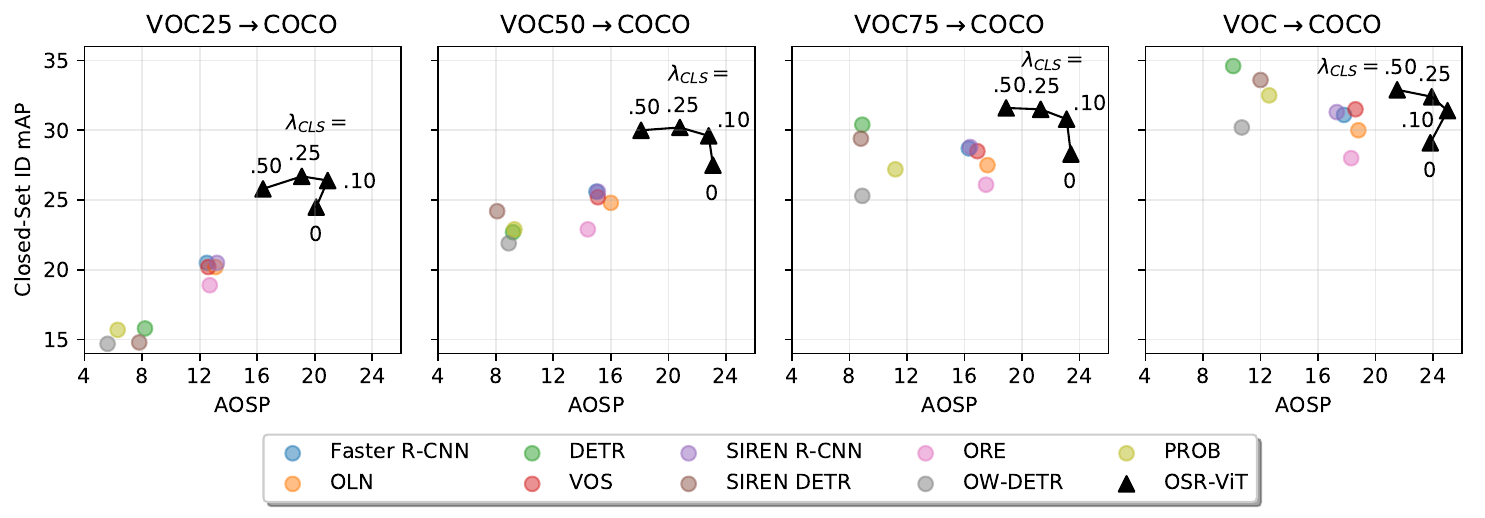}
   \vspace{-3mm}
   \caption{While supervised baselines struggle in the data-constrained settings of our Limited Data Benchmark, our OSR-ViT model maintains good performance.}
   \label{fig:limited_data}
   \vspace{-4mm}
\end{figure}

While performance on large-scale benchmarks is important, in many scenarios and applications we do not have training datasets with hundreds of thousands of annotations at our disposal. For this reason, we devise a Limited Data Benchmark which tests models on semi-supervised versions of the VOC$\rightarrow$COCO task. Here, models are trained on a random (class-balanced) set of 25\%, 50\%, and 75\% of the VOC training annotations and tested on the COCO validation set. 

Fig. \ref{fig:limited_data} visualizes the results from this benchmark as AOSP vs. closed-set ID mAP. Appendix E contains full tabular results. In this experiment we vary the THPN $\lambda_{CLS}$ parameter in our OSR-ViT(-B) model. The key takeaway from this result is that OSR-ViT maintains ID-mAP and AOSP much better than fully supervised models when training data gets scarce. In fact, the most lightweight OSR-ViT model (THPN($\lambda_{CLS}$=.10)+DINOv2-S) trained on 25\% of the VOC data achieves 20.6\% AOSP, which is higher than \textit{any} baseline method trained on 100\% of the VOC data! 
It should also be noted that the CA-AR of the OSR-ViT models trained on the 25\% split is 38.4\%, which essentially matches the highest performing baseline (i.e., OLN) trained on the 100\% split. This tremendous ability is due our framework's ability to realize the full potential of dedicated proposal networks and foundational ViT models for one unified task.

As discussed in Sec. \ref{subsec:natural_imagery_benchmarks} above, some DETR-based baselines outperform our OSR-ViT configuration in terms of closed-set ID-mAP. However, this challenging benchmark reveals that these methods require significant training data to reach this level of performance. Notice that decreasing the labeled training annotations even to 75\% of the original number drastically reduces the performance of these models. In a scenario like VOC25$\rightarrow$COCO, where we have less than 12,000 training annotations, these methods are essentially useless. Finally, these results showcase the effect of THPN's $\lambda_{CLS}$ parameter. In general, the higher we set $\lambda_{CLS}$, the higher the ID mAP. Using an adaptable proposal network like THPN in the OSR-ViT model greatly increases its flexibility, as we can more effectively configure the model for a given set of requirements.

\vspace{-2mm}
\subsection{Ships Benchmark}
\label{subsec:ships_benchmark}
\begin{table}[t]
\caption{Results on the Ships Benchmark.}
\vspace{-3mm}
\centering
\resizebox{0.95\linewidth}{!}{

\begin{tabular}{l|l|llllllll}
\toprule
                                                     &                                  &              &               &               &               & IDvOOD         & IDvNONID       & OODvBG         & FGvBG          \\
Training                                             & Model                            & OOD Algo.    & AOSP        & ID-mAP        & CA-AR         & AUROC          & AUROC          & AUROC          & AUROC          \\ \hline\hline &&&\tabularnewline[-2.2ex]
\multirow{6}{*}{\shortstack[l]{Plain\\ Supervised}} & \multirow{2}{*}{Faster R-CNN}    
& Energy       & 39.7          & 60.8          & 58.6          & 70.55          & 64.65          & 78.00          & 77.38          \\
                                                     &                                  & Mahalanobis  & 40.9          & 60.8          & 58.6          & 45.35          & 55.96          & 78.00          & 77.38          \\ \cline{2-10} &&&\tabularnewline[-2.2ex]
                                                     & \multirow{2}{*}{OLN}             
& Energy       & 46.5          & \textbf{61.3} & 59.7          & 73.17          & 65.99          & 78.59          & 76.14          \\
                                                     &                                  & Mahalanobis  & 46.9          & \textbf{61.3} & 59.7          & 41.66          & 52.07          & 78.59          & 76.14          \\ \cline{2-10} &&&\tabularnewline[-2.2ex]
                                                     & \multirow{2}{*}{Deformable DETR} 
& Energy       & 9.8           & 8.5           & 32.0          & 49.33          & 81.21          & 65.56          & 64.63          \\
                                                     &                                  & Mahalanobis  & 9.5           & 8.5           & 32.0          & 50.70          & 78.45          & 65.56          & 64.63          \\ \hline &&&\tabularnewline[-2.2ex]
VOS                                                  & Faster R-CNN                     & Energy       & 42.6          & 59.5          & 59.0          & 71.49          & 68.45          & 75.05          & 72.70          \\ \hline &&&\tabularnewline[-2.2ex]
\multirow{2}{*}{SIREN}                               & Faster R-CNN                     & SIREN-KNN    & 43.1          & 60.7          & 58.6          & 77.11          & 75.03          & 72.47          & 70.44          \\
                                                     & Deformable DETR                  & SIREN-KNN    & 1.5           & 1.3           & 19.6          & 49.72          & 79.09          & 64.62          & 64.25          \\ \hline &&&\tabularnewline[-2.2ex]
ORE                                                  & Faster R-CNN                     & Energy       & 44.5          & 58.7          & 54.1          & 74.61          & 62.26          & 68.30          & 63.05          \\ \hline &&&\tabularnewline[-2.2ex]
OW-DETR                                              & Deformable DETR                  & Direct Pred. & 8.3           & 11.4          & 31.7          & -              & -              & -              & -              \\ \hline &&&\tabularnewline[-2.2ex]
PROB                                                 & Deformable DETR                  & Direct Pred. & 14.2          & 12.5          & 38.5          & -              & -              & -              & -              \\ \hline &&&\tabularnewline[-2.2ex]
\multirow{2}{*}{OSR-ViT}                                & THPN+DINOv2-S                    & Energy       & 53.4          & 57.2          & \textbf{64.3} & 75.22          & \textbf{87.78} & 94.07          & 95.49          \\
                                                     & THPN+DINOv2-B                    & Energy       & \textbf{55.4} & 58.9 & \textbf{64.3} & \textbf{77.16} & 85.72          & \textbf{94.16} & \textbf{95.81} \\
\bottomrule
\end{tabular}

}
\label{tab:ships}
\end{table}
Our final benchmark evaluates performance in the remote-sensing image domain. We consider the ShipRSImageNet dataset \cite{ShipRSImageNet}, which contains overhead imagery of coastal regions with 50 fine-grained ship classes. Here, we manually create the ID/OOD class split by deeming all ``other'' ship categories as OOD. An implicit challenge of this dataset is that there are relatively few annotations to train on compared to the natural imagery benchmarks (i.e., 2k ship instances compared to 47k VOC instances). Table \ref{tab:ships} contains the results. Even in this different domain, OSR-ViT beats all fully-supervised baselines in terms of AOSP and CA-AR. Our method lags OLN slightly in ID-mAP, but achieves a substantial 8.5\% higher AOSP than OLN's best post-hoc OOD algorithm (Mahalanobis \cite{Mahalanobis}). OSR-ViT's classifier separability is also superior, specifically in terms of objectness-based separability. Our method outperforms the closest baseline (OLN) in OOD vs. BG AUROC and FG vs. BG AUROC by 15.57\% and 19.67\%, respectively! We note that DETR-based methods were unable to converge to a reasonable solution on this smaller-scale task, highlighting their limitations in many settings.

\subsection{OSR-ViT Performance Analysis}
\label{subsec:osrvit_performance_analysis}

\begin{table}[t]
\caption{Model design analysis on the VOC$\rightarrow$COCO task.}
\vspace{-3mm}
\centering
\resizebox{.95\linewidth}{!}{

\begin{tabular}{l|lccccccc}
\toprule
                               &           &               &               &               & IDvOOD         & IDvNONID       & OODvBG         & FGvBG          \\
Model                          & OOD Algo. & AOSP        & ID-mAP        & CA-AR         & AUROC          & AUROC          & AUROC          & AUROC          \\ \hline\hline &&&\tabularnewline[-2.2ex]
\multirow{4}{*}{THPN+DINOv2-B} & MSP       & 24.8          & 31.4          & \textbf{43.2} & 83.97          & 83.33          & 63.42          & 81.86          \\
                               & MaxLogit  & \textbf{25.0}          & 31.4          & \textbf{43.2} & 86.48          & 86.20          & 63.42          & 81.86          \\
                               & ODIN      & \textbf{25.0}          & 31.4          & \textbf{43.2} & 86.00          & 85.41          & 63.42          & 81.86          \\
                               & Energy    & \textbf{25.0}          & 31.4          & \textbf{43.2} & \textbf{86.49}          & \textbf{86.28} & 63.42          & 81.86          \\ \hdashline &&&\tabularnewline[-2.2ex]
FT$\rightarrow$No FT           & Energy    & 24.8          & 31.1          & \textbf{43.2} & 85.46          & 85.41          & 63.87          & \textbf{81.88}          \\
THPN$\rightarrow$Faster R-CNN  & Energy    & 20.0          & \textbf{32.4}          & 37.2          & 84.38          & 82.51          & 61.59          & 71.81          \\
DINOv2-B$\rightarrow$CLIP-B    & Energy    & 22.3          & 29.0          & \textbf{43.2} & 78.74          & 83.53          & \textbf{64.70}          & 80.21         \\
\bottomrule
\end{tabular}

}
\label{tab:model_design_analysis}
\end{table}

OSR-ViT's modular design allows for arbitrary proposal networks and feature extractors to be incorporated. In Table \ref{tab:model_design_analysis} we investigate several different variants of our base configuration using THPN and DINOv2-B on the VOC$\rightarrow$COCO task. The exact choice of post-hoc OOD algorithm does not have a massive effect on performance, although Energy is the best overall. The FT$\rightarrow$No FT row represents our base configuration but without the 5-epoch end-to-end fine-tuning step described in Sec. \ref{subsec:foundational_classifier}. While this fine-tuning is not necessary, it does boost overall performance. When we swap THPN ($\lambda_{CLS}$=.10) for a class-agnostic Faster R-CNN \cite{FasterRCNN} proposal network, we get noticeably worse AOSP and CA-AR, but better ID-mAP due to Faster R-CNN's inherent ID bias. But again, it should be noted that a THPN with $\lambda_{CLS}$=.50 can outperform Faster R-CNN with an ID mAP of 32.9. Finally, we compare the impact of swapping the DINOv2 foundational model for a CLIP \cite{CLIP} model of the same size. We find that OSR-ViT with CLIP achieves substandard results across the board.

\begin{figure}[t]
  \centering
   \includegraphics[width=0.95\linewidth]{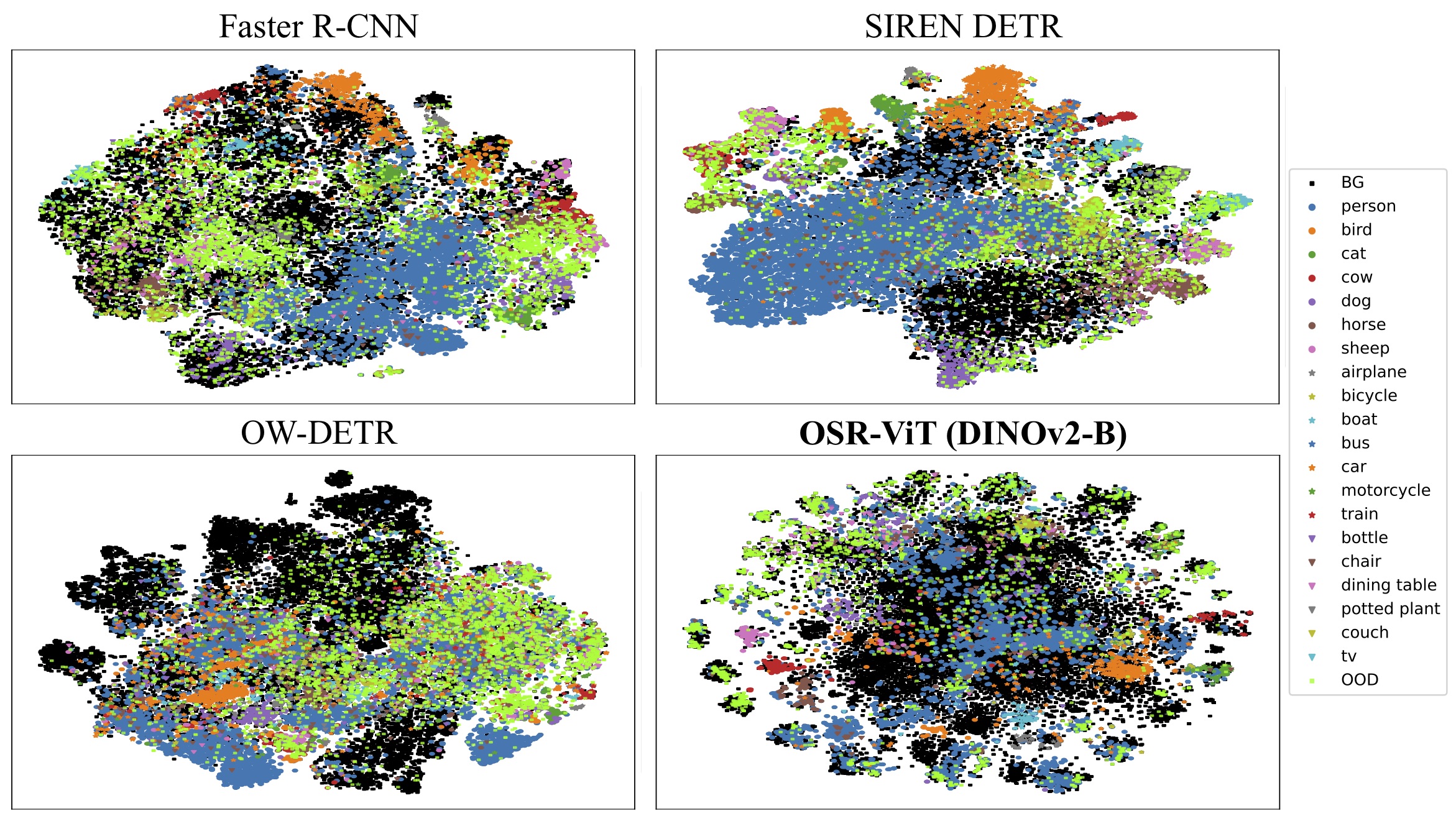}
   \vspace{-3mm}
   \caption{2D t-SNE visualization of penultimate features on the VOC$\rightarrow$COCO task. OSR-ViT models generate the most compact ID-class clusters, aiding in ID vs. OOD separation. Also, OSR-ViT's ability to segregate OOD instances into different (likely class-wise) clusters is extremely difficult to emulate with only task-specific supervision.}
   \label{fig:tsne}
   \vspace{-3mm}
\end{figure}
Figure \ref{fig:tsne} depicts 2D t-SNE visualizations \cite{tSNE} of the penultimate object features of four different models on the VOC$\rightarrow$COCO task. Note that the colored circle, star, and triangle markers represent detections that positively match ID ground-truth objects, the chartreuse squares represent detections matched to OOD ground truth objects, and the black squares represent detections matched to background. Ultimately, the performance of a model is directly related to how separable these features are, with more compact ID and OOD clusters being indicative of better models. The key takeaway from this analysis is that the DINOv2 feature extractor does a far better job of separating the OOD objects from ID objects and BG compared to the baselines. These findings support our hypothesis from Sec. \ref{subsec:foundational_classifier}: The DINOv2 representations are indeed nuanced enough to not only distinguish ID and OOD objects, but also different OOD objects from each other. This quality of representation is generally not feasible with task-specific supervised training alone. Finally, this OOD separability would make our method a powerful starting starting point for the OWOD task which incrementally learns new classes, but we leave this for future work. An additional analysis on model size and visualizated samples are in the Appendix.
\vspace{-2mm}

\section{Conclusion}
\label{sec:conclusion}

As ML becomes more and more ubiquitous in our real-world systems, it is important to keep safety at the forefront of model design. In this work, we identify a serious vulnerability of state-of-the-art ``open-set object detection'' models: the detection of unknown objects is not explicitly prioritized. We use this finding to motivate a new OSODD task, and create a comprehensive evaluation protocol that allows different related works to be directly compared to each other for the first time. We also introduce a modular bipartite OSR-ViT framework that leverages a self-contained proposal network and and off-the-shelf ViT model to achieve far superior performance to supervised baselines. Our OSR-ViT approach embraces the recent push towards using internet-scale foundational models as the basis for task-specific applications by allowing for the integration of such models in a plug-and-play fashion. We argue that this style of solution is not only more powerful but also more future-proof, as it prevents rapid obsolescence.



%
%
\bibliographystyle{splncs04}
\bibliography{egbib}

\begin{thebibliography}{10}
\providecommand{\url}[1]{\texttt{#1}}
\providecommand{\urlprefix}{URL }
\providecommand{\doi}[1]{https://doi.org/#1}

\bibitem{DissimilarityCoeffBasedObjDiscovery}
Arun, A., Jawahar, C.V., Kumar, M.P.: Dissimilarity coefficient based weakly supervised object detection. In: {IEEE} Conference on Computer Vision and Pattern Recognition, {CVPR} 2019, Long Beach, CA, USA, June 16-20, 2019. pp. 9432--9441. Computer Vision Foundation / {IEEE} (2019). \doi{10.1109/CVPR.2019.00966}, \url{http://openaccess.thecvf.com/content\_CVPR\_2019/html/Arun\_Dissimilarity\_Coefficient\_Based\_Weakly\_Supervised\_Object\_Detection\_CVPR\_2019\_paper.html}

\bibitem{DETR}
Carion, N., Massa, F., Synnaeve, G., Usunier, N., Kirillov, A., Zagoruyko, S.: End-to-end object detection with transformers. In: Vedaldi, A., Bischof, H., Brox, T., Frahm, J. (eds.) Computer Vision - {ECCV} 2020 - 16th European Conference, Glasgow, UK, August 23-28, 2020, Proceedings, Part {I}. Lecture Notes in Computer Science, vol. 12346, pp. 213--229. Springer (2020). \doi{10.1007/978-3-030-58452-8\_13}, \url{https://doi.org/10.1007/978-3-030-58452-8\_13}

\bibitem{ObjectDiscoveryInTheWild}
Cho, M., Kwak, S., Schmid, C., Ponce, J.: Unsupervised object discovery and localization in the wild: Part-based matching with bottom-up region proposals. In: {IEEE} Conference on Computer Vision and Pattern Recognition, {CVPR} 2015, Boston, MA, USA, June 7-12, 2015. pp. 1201--1210. {IEEE} Computer Society (2015). \doi{10.1109/CVPR.2015.7298724}, \url{https://doi.org/10.1109/CVPR.2015.7298724}

\bibitem{deng2024medical}
Deng, H., Zhang, Z., Bao, J., Li, X.: Medical open set recognition via intra-class clustering. In: Medical Imaging with Deep Learning (2024)

\bibitem{OverlookedElephantOpenSet}
Dhamija, A.R., G{\"{u}}nther, M., Ventura, J., Boult, T.E.: The overlooked elephant of object detection: Open set. In: {IEEE} Winter Conference on Applications of Computer Vision, {WACV} 2020, Snowmass Village, CO, USA, March 1-5, 2020. pp. 1010--1019. {IEEE} (2020). \doi{10.1109/WACV45572.2020.9093355}, \url{https://doi.org/10.1109/WACV45572.2020.9093355}

\bibitem{ViT}
Dosovitskiy, A., Beyer, L., Kolesnikov, A., Weissenborn, D., Zhai, X., Unterthiner, T., Dehghani, M., Minderer, M., Heigold, G., Gelly, S., Uszkoreit, J., Houlsby, N.: An image is worth 16x16 words: Transformers for image recognition at scale. In: 9th International Conference on Learning Representations, {ICLR} 2021, Virtual Event, Austria, May 3-7, 2021. OpenReview.net (2021), \url{https://openreview.net/forum?id=YicbFdNTTy}

\bibitem{SIREN}
Du, X., Gozum, G., Ming, Y., Li, Y.: {SIREN:} shaping representations for detecting out-of-distribution objects. In: Koyejo, S., Mohamed, S., Agarwal, A., Belgrave, D., Cho, K., Oh, A. (eds.) Advances in Neural Information Processing Systems 35: Annual Conference on Neural Information Processing Systems 2022, NeurIPS 2022, New Orleans, LA, USA, November 28 - December 9, 2022 (2022), \url{http://papers.nips.cc/paper\_files/paper/2022/hash/804dbf8d3b8eee1ef875c6857efc64eb-Abstract-Conference.html}

\bibitem{STUD}
Du, X., Wang, X., Gozum, G., Li, Y.: Unknown-aware object detection: Learning what you don't know from videos in the wild. In: {IEEE/CVF} Conference on Computer Vision and Pattern Recognition, {CVPR} 2022, New Orleans, LA, USA, June 18-24, 2022. pp. 13668--13678. {IEEE} (2022). \doi{10.1109/CVPR52688.2022.01331}, \url{https://doi.org/10.1109/CVPR52688.2022.01331}

\bibitem{VOS}
Du, X., Wang, Z., Cai, M., Li, Y.: {VOS:} learning what you don't know by virtual outlier synthesis. CoRR  \textbf{abs/2202.01197} (2022), \url{https://arxiv.org/abs/2202.01197}

\bibitem{ObjectDiscoveryRGBD}
Ekekrantz, J., Bore, N., Ambrus, R., Folkesson, J., Jensfelt, P.: Unsupervised object discovery and segmentation of rgbd-images. CoRR  \textbf{abs/1710.06929} (2017), \url{http://arxiv.org/abs/1710.06929}

\bibitem{PASCALVOC}
Everingham, M., Gool, L.V., Williams, C.K.I., Winn, J.M., Zisserman, A.: The pascal visual object classes {(VOC)} challenge. Int. J. Comput. Vis.  \textbf{88}(2),  303--338 (2010). \doi{10.1007/s11263-009-0275-4}, \url{https://doi.org/10.1007/s11263-009-0275-4}

\bibitem{DropoutSampling}
Gal, Y., Ghahramani, Z.: Dropout as a bayesian approximation: Representing model uncertainty in deep learning. In: Balcan, M., Weinberger, K.Q. (eds.) Proceedings of the 33nd International Conference on Machine Learning, {ICML} 2016, New York City, NY, USA, June 19-24, 2016. {JMLR} Workshop and Conference Proceedings, vol.~48, pp. 1050--1059. JMLR.org (2016), \url{http://proceedings.mlr.press/v48/gal16.html}

\bibitem{FastRCNN}
Girshick, R.B.: Fast {R-CNN}. In: 2015 {IEEE} International Conference on Computer Vision, {ICCV} 2015, Santiago, Chile, December 7-13, 2015. pp. 1440--1448. {IEEE} Computer Society (2015). \doi{10.1109/ICCV.2015.169}, \url{https://doi.org/10.1109/ICCV.2015.169}

\bibitem{RCNN}
Girshick, R.B., Donahue, J., Darrell, T., Malik, J.: Rich feature hierarchies for accurate object detection and semantic segmentation. In: 2014 {IEEE} Conference on Computer Vision and Pattern Recognition, {CVPR} 2014, Columbus, OH, USA, June 23-28, 2014. pp. 580--587. {IEEE} Computer Society (2014). \doi{10.1109/CVPR.2014.81}, \url{https://doi.org/10.1109/CVPR.2014.81}

\bibitem{OWDETR}
Gupta, A., Narayan, S., Joseph, K.J., Khan, S., Khan, F.S., Shah, M.: {OW-DETR:} open-world detection transformer. In: {IEEE/CVF} Conference on Computer Vision and Pattern Recognition, {CVPR} 2022, New Orleans, LA, USA, June 18-24, 2022. pp. 9225--9234. {IEEE} (2022). \doi{10.1109/CVPR52688.2022.00902}, \url{https://doi.org/10.1109/CVPR52688.2022.00902}

\bibitem{ExpandingLowDensityLatentRegionsOSOD}
Han, J., Ren, Y., Ding, J., Pan, X., Yan, K., Xia, G.: Expanding low-density latent regions for open-set object detection. CoRR  \textbf{abs/2203.14911} (2022). \doi{10.48550/arXiv.2203.14911}, \url{https://doi.org/10.48550/arXiv.2203.14911}

\bibitem{MaskRCNN}
He, K., Gkioxari, G., Doll{\'{a}}r, P., Girshick, R.B.: Mask {R-CNN}. In: {IEEE} International Conference on Computer Vision, {ICCV} 2017, Venice, Italy, October 22-29, 2017. pp. 2980--2988. {IEEE} Computer Society (2017). \doi{10.1109/ICCV.2017.322}, \url{https://doi.org/10.1109/ICCV.2017.322}

\bibitem{ResNet}
He, K., Zhang, X., Ren, S., Sun, J.: Deep residual learning for image recognition. In: 2016 {IEEE} Conference on Computer Vision and Pattern Recognition, {CVPR} 2016, Las Vegas, NV, USA, June 27-30, 2016. pp. 770--778. {IEEE} Computer Society (2016). \doi{10.1109/CVPR.2016.90}, \url{https://doi.org/10.1109/CVPR.2016.90}

\bibitem{MSP}
Hendrycks, D., Gimpel, K.: A baseline for detecting misclassified and out-of-distribution examples in neural networks. In: 5th International Conference on Learning Representations, {ICLR} 2017, Toulon, France, April 24-26, 2017, Conference Track Proceedings. OpenReview.net (2017), \url{https://openreview.net/forum?id=Hkg4TI9xl}

\bibitem{huang2019apolloscape}
Huang, X., Wang, P., Cheng, X., Zhou, D., Geng, Q., Yang, R.: The apolloscape open dataset for autonomous driving and its application. IEEE transactions on pattern analysis and machine intelligence  \textbf{42}(10),  2702--2719 (2019)

\bibitem{THPN}
Inkawhich, M., Inkawhich, N., Li, H., Chen, Y.: Tunable hybrid proposal networks for the open world. In: {IEEE} Winter Conference on Applications of Computer Vision, {WACV} 2024, Waikoloa Beach, HI, USA, January 4-8, 2024. pp. 1988--1999. {IEEE} (2024)

\bibitem{BatchNormalization}
Ioffe, S., Szegedy, C.: Batch normalization: Accelerating deep network training by reducing internal covariate shift. In: Bach, F.R., Blei, D.M. (eds.) Proceedings of the 32nd International Conference on Machine Learning, {ICML} 2015, Lille, France, 6-11 July 2015. {JMLR} Workshop and Conference Proceedings, vol.~37, pp. 448--456. JMLR.org (2015), \url{http://proceedings.mlr.press/v37/ioffe15.html}

\bibitem{TowardsOpenWorldObjectDetection}
Joseph, K.J., Khan, S.H., Khan, F.S., Balasubramanian, V.N.: Towards open world object detection. In: {IEEE} Conference on Computer Vision and Pattern Recognition, {CVPR} 2021, virtual, June 19-25, 2021. pp. 5830--5840. Computer Vision Foundation / {IEEE} (2021), \url{https://openaccess.thecvf.com/content/CVPR2021/html/Joseph\_Towards\_Open\_World\_Object\_Detection\_CVPR\_2021\_paper.html}

\bibitem{OLN}
Kim, D., Lin, T., Angelova, A., Kweon, I.S., Kuo, W.: Learning open-world object proposals without learning to classify. {IEEE} Robotics Autom. Lett.  \textbf{7}(2),  5453--5460 (2022). \doi{10.1109/LRA.2022.3146922}, \url{https://doi.org/10.1109/LRA.2022.3146922}

\bibitem{KevinsPaper}
Konan, S., Liang, K.J., Yin, L.: Extending one-stage detection with open-world proposals. CoRR  \textbf{abs/2201.02302} (2022), \url{https://arxiv.org/abs/2201.02302}

\bibitem{Lee_pseudolabel}
hyun Lee, D.: Pseudo-label: The simple and efficient semi-supervised learning method for deep neural networks (2013)

\bibitem{Mahalanobis}
Lee, K., Lee, K., Lee, H., Shin, J.: A simple unified framework for detecting out-of-distribution samples and adversarial attacks. In: Bengio, S., Wallach, H.M., Larochelle, H., Grauman, K., Cesa{-}Bianchi, N., Garnett, R. (eds.) Advances in Neural Information Processing Systems 31: Annual Conference on Neural Information Processing Systems 2018, NeurIPS 2018, December 3-8, 2018, Montr{\'{e}}al, Canada. pp. 7167--7177 (2018), \url{https://proceedings.neurips.cc/paper/2018/hash/abdeb6f575ac5c6676b747bca8d09cc2-Abstract.html}

\bibitem{FPN}
Lin, T., Doll{\'{a}}r, P., Girshick, R.B., He, K., Hariharan, B., Belongie, S.J.: Feature pyramid networks for object detection. In: 2017 {IEEE} Conference on Computer Vision and Pattern Recognition, {CVPR} 2017, Honolulu, HI, USA, July 21-26, 2017. pp. 936--944. {IEEE} Computer Society (2017). \doi{10.1109/CVPR.2017.106}, \url{https://doi.org/10.1109/CVPR.2017.106}

\bibitem{COCO}
Lin, T., Maire, M., Belongie, S.J., Hays, J., Perona, P., Ramanan, D., Doll{\'{a}}r, P., Zitnick, C.L.: Microsoft {COCO:} common objects in context. In: Fleet, D.J., Pajdla, T., Schiele, B., Tuytelaars, T. (eds.) Computer Vision - {ECCV} 2014 - 13th European Conference, Zurich, Switzerland, September 6-12, 2014, Proceedings, Part {V}. Lecture Notes in Computer Science, vol.~8693, pp. 740--755. Springer (2014). \doi{10.1007/978-3-319-10602-1\_48}, \url{https://doi.org/10.1007/978-3-319-10602-1\_48}

\bibitem{SSD}
Liu, W., Anguelov, D., Erhan, D., Szegedy, C., Reed, S.E., Fu, C., Berg, A.C.: {SSD:} single shot multibox detector. In: Leibe, B., Matas, J., Sebe, N., Welling, M. (eds.) Computer Vision - {ECCV} 2016 - 14th European Conference, Amsterdam, The Netherlands, October 11-14, 2016, Proceedings, Part {I}. Lecture Notes in Computer Science, vol.~9905, pp. 21--37. Springer (2016). \doi{10.1007/978-3-319-46448-0\_2}, \url{https://doi.org/10.1007/978-3-319-46448-0\_2}

\bibitem{EnergyOOD}
Liu, W., Wang, X., Owens, J.D., Li, Y.: Energy-based out-of-distribution detection. In: Larochelle, H., Ranzato, M., Hadsell, R., Balcan, M., Lin, H. (eds.) Advances in Neural Information Processing Systems 33: Annual Conference on Neural Information Processing Systems 2020, NeurIPS 2020, December 6-12, 2020, virtual (2020), \url{https://proceedings.neurips.cc/paper/2020/hash/f5496252609c43eb8a3d147ab9b9c006-Abstract.html}

\bibitem{ObjectCentricLearningSlotAttention}
Locatello, F., Weissenborn, D., Unterthiner, T., Mahendran, A., Heigold, G., Uszkoreit, J., Dosovitskiy, A., Kipf, T.: Object-centric learning with slot attention. In: Larochelle, H., Ranzato, M., Hadsell, R., Balcan, M., Lin, H. (eds.) Advances in Neural Information Processing Systems 33: Annual Conference on Neural Information Processing Systems 2020, NeurIPS 2020, December 6-12, 2020, virtual (2020), \url{https://proceedings.neurips.cc/paper/2020/hash/8511df98c02ab60aea1b2356c013bc0f-Abstract.html}

\bibitem{tSNE}
van~der Maaten, L., Hinton, G.: Visualizing data using t-sne. Journal of Machine Learning Research  \textbf{9}(86),  2579--2605 (2008), \url{http://jmlr.org/papers/v9/vandermaaten08a.html}

\bibitem{Miller2}
Miller, D., Dayoub, F., Milford, M., S{\"{u}}nderhauf, N.: Evaluating merging strategies for sampling-based uncertainty techniques in object detection. In: International Conference on Robotics and Automation, {ICRA} 2019, Montreal, QC, Canada, May 20-24, 2019. pp. 2348--2354. {IEEE} (2019). \doi{10.1109/ICRA.2019.8793821}, \url{https://doi.org/10.1109/ICRA.2019.8793821}

\bibitem{Miller1}
Miller, D., Nicholson, L., Dayoub, F., S{\"{u}}nderhauf, N.: Dropout sampling for robust object detection in open-set conditions. In: 2018 {IEEE} International Conference on Robotics and Automation, {ICRA} 2018, Brisbane, Australia, May 21-25, 2018. pp.~1--7. {IEEE} (2018). \doi{10.1109/ICRA.2018.8460700}, \url{https://doi.org/10.1109/ICRA.2018.8460700}

\bibitem{Miller3}
Miller, D., S{\"{u}}nderhauf, N., Milford, M., Dayoub, F.: Uncertainty for identifying open-set errors in visual object detection. {IEEE} Robotics Autom. Lett.  \textbf{7}(1),  215--222 (2022). \doi{10.1109/LRA.2021.3123374}, \url{https://doi.org/10.1109/LRA.2021.3123374}

\bibitem{ScalingOVOD}
Minderer, M., Gritsenko, A.A., Houlsby, N.: Scaling open-vocabulary object detection. CoRR  \textbf{abs/2306.09683} (2023). \doi{10.48550/ARXIV.2306.09683}, \url{https://doi.org/10.48550/arXiv.2306.09683}

\bibitem{moon2022difficulty}
Moon, W., Park, J., Seong, H.S., Cho, C.H., Heo, J.P.: Difficulty-aware simulator for open set recognition. In: European Conference on Computer Vision. pp. 365--381. Springer (2022)

\bibitem{DINOv2}
Oquab, M., Darcet, T., Moutakanni, T., Vo, H., Szafraniec, M., Khalidov, V., Fernandez, P., Haziza, D., Massa, F., El{-}Nouby, A., Assran, M., Ballas, N., Galuba, W., Howes, R., Huang, P., Li, S., Misra, I., Rabbat, M.G., Sharma, V., Synnaeve, G., Xu, H., J{\'{e}}gou, H., Mairal, J., Labatut, P., Joulin, A., Bojanowski, P.: Dinov2: Learning robust visual features without supervision. CoRR  \textbf{abs/2304.07193} (2023). \doi{10.48550/ARXIV.2304.07193}, \url{https://doi.org/10.48550/arXiv.2304.07193}

\bibitem{CLIP}
Radford, A., Kim, J.W., Hallacy, C., Ramesh, A., Goh, G., Agarwal, S., Sastry, G., Askell, A., Mishkin, P., Clark, J., Krueger, G., Sutskever, I.: Learning transferable visual models from natural language supervision. In: Meila, M., Zhang, T. (eds.) Proceedings of the 38th International Conference on Machine Learning, {ICML} 2021, 18-24 July 2021, Virtual Event. Proceedings of Machine Learning Research, vol.~139, pp. 8748--8763. {PMLR} (2021), \url{http://proceedings.mlr.press/v139/radford21a.html}

\bibitem{ObjectCentricOVD}
Rasheed, H.A., Maaz, M., Khattak, M.U., Khan, S.H., Khan, F.S.: Bridging the gap between object and image-level representations for open-vocabulary detection. In: Koyejo, S., Mohamed, S., Agarwal, A., Belgrave, D., Cho, K., Oh, A. (eds.) Advances in Neural Information Processing Systems 35: Annual Conference on Neural Information Processing Systems 2022, NeurIPS 2022, New Orleans, LA, USA, November 28 - December 9, 2022 (2022), \url{http://papers.nips.cc/paper\_files/paper/2022/hash/dabf612543b97ea9c8f46d058d33cf74-Abstract-Conference.html}

\bibitem{YOLO}
Redmon, J., Divvala, S.K., Girshick, R.B., Farhadi, A.: You only look once: Unified, real-time object detection. In: 2016 {IEEE} Conference on Computer Vision and Pattern Recognition, {CVPR} 2016, Las Vegas, NV, USA, June 27-30, 2016. pp. 779--788. {IEEE} Computer Society (2016). \doi{10.1109/CVPR.2016.91}, \url{https://doi.org/10.1109/CVPR.2016.91}

\bibitem{FasterRCNN}
Ren, S., He, K., Girshick, R.B., Sun, J.: Faster {R-CNN:} towards real-time object detection with region proposal networks. In: Cortes, C., Lawrence, N.D., Lee, D.D., Sugiyama, M., Garnett, R. (eds.) Advances in Neural Information Processing Systems 28: Annual Conference on Neural Information Processing Systems 2015, December 7-12, 2015, Montreal, Quebec, Canada. pp. 91--99 (2015), \url{https://proceedings.neurips.cc/paper/2015/hash/14bfa6bb14875e45bba028a21ed38046-Abstract.html}

\bibitem{OGObjectDiscovery}
Rubinstein, M., Joulin, A., Kopf, J., Liu, C.: Unsupervised joint object discovery and segmentation in internet images. In: 2013 {IEEE} Conference on Computer Vision and Pattern Recognition, Portland, OR, USA, June 23-28, 2013. pp. 1939--1946. {IEEE} Computer Society (2013). \doi{10.1109/CVPR.2013.253}, \url{https://doi.org/10.1109/CVPR.2013.253}

\bibitem{LearningToDetectEveryThing}
Saito, K., Hu, P., Darrell, T., Saenko, K.: Learning to detect every thing in an open world. CoRR  \textbf{abs/2112.01698} (2021), \url{https://arxiv.org/abs/2112.01698}

\bibitem{OGOpenSet}
Scheirer, W.J., de~Rezende~Rocha, A., Sapkota, A., Boult, T.E.: Toward open set recognition. {IEEE} Trans. Pattern Anal. Mach. Intell.  \textbf{35}(7),  1757--1772 (2013). \doi{10.1109/TPAMI.2012.256}, \url{https://doi.org/10.1109/TPAMI.2012.256}

\bibitem{Objects365}
Shao, S., Li, Z., Zhang, T., Peng, C., Yu, G., Zhang, X., Li, J., Sun, J.: Objects365: {A} large-scale, high-quality dataset for object detection. In: 2019 {IEEE/CVF} International Conference on Computer Vision, {ICCV} 2019, Seoul, Korea (South), October 27 - November 2, 2019. pp. 8429--8438. {IEEE} (2019). \doi{10.1109/ICCV.2019.00852}, \url{https://doi.org/10.1109/ICCV.2019.00852}

\bibitem{sun2020scalability}
Sun, P., Kretzschmar, H., Dotiwalla, X., Chouard, A., Patnaik, V., Tsui, P., Guo, J., Zhou, Y., Chai, Y., Caine, B., et~al.: Scalability in perception for autonomous driving: Waymo open dataset. In: Proceedings of the IEEE/CVF conference on computer vision and pattern recognition. pp. 2446--2454 (2020)

\bibitem{CascadeRPN}
Vu, T., Jang, H., Pham, T.X., Yoo, C.D.: Cascade {RPN:} delving into high-quality region proposal network with adaptive convolution. In: Wallach, H.M., Larochelle, H., Beygelzimer, A., d'Alch{\'{e}}{-}Buc, F., Fox, E.B., Garnett, R. (eds.) Advances in Neural Information Processing Systems 32: Annual Conference on Neural Information Processing Systems 2019, NeurIPS 2019, December 8-14, 2019, Vancouver, BC, Canada. pp. 1430--1440 (2019), \url{https://proceedings.neurips.cc/paper/2019/hash/d554f7bb7be44a7267068a7df88ddd20-Abstract.html}

\bibitem{RegionProposalGuidedAnchoring}
Wang, J., Chen, K., Yang, S., Loy, C.C., Lin, D.: Region proposal by guided anchoring. In: {IEEE} Conference on Computer Vision and Pattern Recognition, {CVPR} 2019, Long Beach, CA, USA, June 16-20, 2019. pp. 2965--2974. Computer Vision Foundation / {IEEE} (2019). \doi{10.1109/CVPR.2019.00308}, \url{http://openaccess.thecvf.com/content\_CVPR\_2019/html/Wang\_Region\_Proposal\_by\_Guided\_Anchoring\_CVPR\_2019\_paper.html}

\bibitem{CORA}
Wu, X., Zhu, F., Zhao, R., Li, H.: {CORA:} adapting {CLIP} for open-vocabulary detection with region prompting and anchor pre-matching. In: {IEEE/CVF} Conference on Computer Vision and Pattern Recognition, {CVPR} 2023, Vancouver, BC, Canada, June 17-24, 2023. pp. 7031--7040. {IEEE} (2023). \doi{10.1109/CVPR52729.2023.00679}, \url{https://doi.org/10.1109/CVPR52729.2023.00679}

\bibitem{TwoBranchObjCentricOWOD}
Wu, Y., Zhao, X., Ma, Y., Wang, D., Liu, X.: Two-branch objectness-centric open world detection. In: Zhang, D., Fang, C., Liu, W., Liu, X., Song, J., Zhu, H., Huang, W., Smith, J. (eds.) HCMA@MM 2022: Proceedings of the 3rd International Workshop on Human-Centric Multimedia Analysis, Lisboa, Portugal, October 10, 2022. pp. 35--40. {ACM} (2022). \doi{10.1145/3552458.3556453}, \url{https://doi.org/10.1145/3552458.3556453}

\bibitem{UCOWOD}
Wu, Z., Lu, Y., Chen, X., Wu, Z., Kang, L., Yu, J.: {UC-OWOD:} unknown-classified open world object detection. In: Avidan, S., Brostow, G.J., Ciss{\'{e}}, M., Farinella, G.M., Hassner, T. (eds.) Computer Vision - {ECCV} 2022 - 17th European Conference, Tel Aviv, Israel, October 23-27, 2022, Proceedings, Part {X}. Lecture Notes in Computer Science, vol. 13670, pp. 193--210. Springer (2022). \doi{10.1007/978-3-031-20080-9\_12}, \url{https://doi.org/10.1007/978-3-031-20080-9\_12}

\bibitem{ObjectsInSemanticTopology}
Yang, S., Sun, P., Jiang, Y., Xia, X., Zhang, R., Yuan, Z., Wang, C., Luo, P., Xu, M.: Objects in semantic topology. In: The Tenth International Conference on Learning Representations, {ICLR} 2022, Virtual Event, April 25-29, 2022. OpenReview.net (2022), \url{https://openreview.net/forum?id=d5SCUJ5t1k}

\bibitem{OWODDiscriminativeClassPrototype}
Yu, J., Ma, L., Li, Z., Peng, Y., Xie, S.: Open-world object detection via discriminative class prototype learning. In: 2022 {IEEE} International Conference on Image Processing, {ICIP} 2022, Bordeaux, France, 16-19 October 2022. pp. 626--630. {IEEE} (2022). \doi{10.1109/ICIP46576.2022.9897461}, \url{https://doi.org/10.1109/ICIP46576.2022.9897461}

\bibitem{OVDETR}
Zang, Y., Li, W., Zhou, K., Huang, C., Loy, C.C.: Open-vocabulary {DETR} with conditional matching. In: Avidan, S., Brostow, G.J., Ciss{\'{e}}, M., Farinella, G.M., Hassner, T. (eds.) Computer Vision - {ECCV} 2022 - 17th European Conference, Tel Aviv, Israel, October 23-27, 2022, Proceedings, Part {IX}. Lecture Notes in Computer Science, vol. 13669, pp. 106--122. Springer (2022). \doi{10.1007/978-3-031-20077-9\_7}, \url{https://doi.org/10.1007/978-3-031-20077-9\_7}

\bibitem{OVODusingCaptions}
Zareian, A., Rosa, K.D., Hu, D.H., Chang, S.: Open-vocabulary object detection using captions. In: {IEEE} Conference on Computer Vision and Pattern Recognition, {CVPR} 2021, virtual, June 19-25, 2021. pp. 14393--14402. Computer Vision Foundation / {IEEE} (2021). \doi{10.1109/CVPR46437.2021.01416}, \url{https://openaccess.thecvf.com/content/CVPR2021/html/Zareian\_Open-Vocabulary\_Object\_Detection\_Using\_Captions\_CVPR\_2021\_paper.html}

\bibitem{ShipRSImageNet}
Zhang, Z., Zhang, L., Wang, Y., Feng, P., He, R.: Shiprsimagenet: {A} large-scale fine-grained dataset for ship detection in high-resolution optical remote sensing images. {IEEE} J. Sel. Top. Appl. Earth Obs. Remote. Sens.  \textbf{14},  8458--8472 (2021). \doi{10.1109/JSTARS.2021.3104230}, \url{https://doi.org/10.1109/JSTARS.2021.3104230}

\bibitem{RevisitingOWOD}
Zhao, X., Liu, X., Shen, Y., Qiao, Y., Ma, Y., Wang, D.: Revisiting open world object detection. CoRR  \textbf{abs/2201.00471} (2022), \url{https://arxiv.org/abs/2201.00471}

\bibitem{DeformableDETR}
Zhu, X., Su, W., Lu, L., Li, B., Wang, X., Dai, J.: Deformable {DETR:} deformable transformers for end-to-end object detection. In: 9th International Conference on Learning Representations, {ICLR} 2021, Virtual Event, Austria, May 3-7, 2021. OpenReview.net (2021), \url{https://openreview.net/forum?id=gZ9hCDWe6ke}

\bibitem{PROB}
Zohar, O., Wang, K., Yeung, S.: {PROB:} probabilistic objectness for open world object detection. In: {IEEE/CVF} Conference on Computer Vision and Pattern Recognition, {CVPR} 2023, Vancouver, BC, Canada, June 17-24, 2023. pp. 11444--11453. {IEEE} (2023). \doi{10.1109/CVPR52729.2023.01101}, \url{https://doi.org/10.1109/CVPR52729.2023.01101}

\end{thebibliography}

\clearpage
\section*{Appendix}

\subsection*{A. Implementation Details}
\textbf{Prediction Partitioning.}
In order to compute the AUROC metrics without requiring mutually exclusive ID/OOD validation sets, we use a prediction partitioning operation (See Sec. \ref{subsec:osodd_evaluation_protocol}). This operation matches each prediction to its corresponding ID/OOD/BG bin based on its IoU overlap with ground truth boxes from a mixed test set (i.e., the images contain both ID and OOD objects). 
Predictions with an intersection-over-union (IoU) $\ge$ 0.5 with some ground truth box is matched to the corresponding ground truth, and predictions with an IoU $<$ 0.2 with any ground truth box is deemed a background match. Predictions with 0.2 $\le$ IoU $<$ 0.5 are ignored. 
Note that during evaluation, we always pretend that some subset of classes are OOD, so we have ground truth matches for OOD objects too. 

\textbf{Proposal Network.}
We use the default configuration of THPN \cite{THPN} for most of this work. Our THPN uses a ResNet-50 \cite{ResNet} with a Feature Pyramid Network (FPN) \cite{FPN} as a backbone network. We train on the initial labeled training set for 16 epochs, then perform 2 stages of self-training which adds 30\% more pseudo-labels compared to the original set. During inference, the THPN outputs 100 proposals per image.

\textbf{Foundational Classifier.}
In the Region Extract operation we crop a 10\% larger area than the predicted proposal box to capture slightly more surrounding contextual information. When resizing the cropped regions, we first resize the longer side to 224, then zero pad the shorter side to 224 so that the aspect ratio of the region is preserved. Our foundational classifier consists of an off-the-shelf DINOv2 model \cite{DINOv2} from the official repository, followed by a 2-layer fully-connected head $f$ to facilitate the C-way classification. This classification head includes a ReLU activation for non-linearity and a batch-normalization \cite{BatchNormalization} operation. To train, we use standard cross-entropy loss, an SGD optimizer, and minority-class oversampling to handle data imbalance. As mentioned in Sec. \ref{subsec:training}, we train this foundational classifier in two stages. First, we freeze the DINOv2-pretrained ViT $V$ and only update $f$ for 50 epochs using a learning rate of 0.1. Here, we use a batch size of 256 and decay the learning rate by 10x at epochs 20 and 35. Next, we perform a short 5-epoch fine-tuning stage in which we train both $V$ and $f$ together with a batch size of 42 (to fit on the device) and a learning rate of 0.000001. In both stages, we set SGD momentum to 0.9 and use $L_2$ regularization with strength 0.0001. We use a single V100 GPU to train each foundational classifier.

We will open source our code upon acceptance.

\subsection*{B. Classifier Separability Analysis}
\begin{figure}[t]
  \centering
   \includegraphics[width=1.0\linewidth]{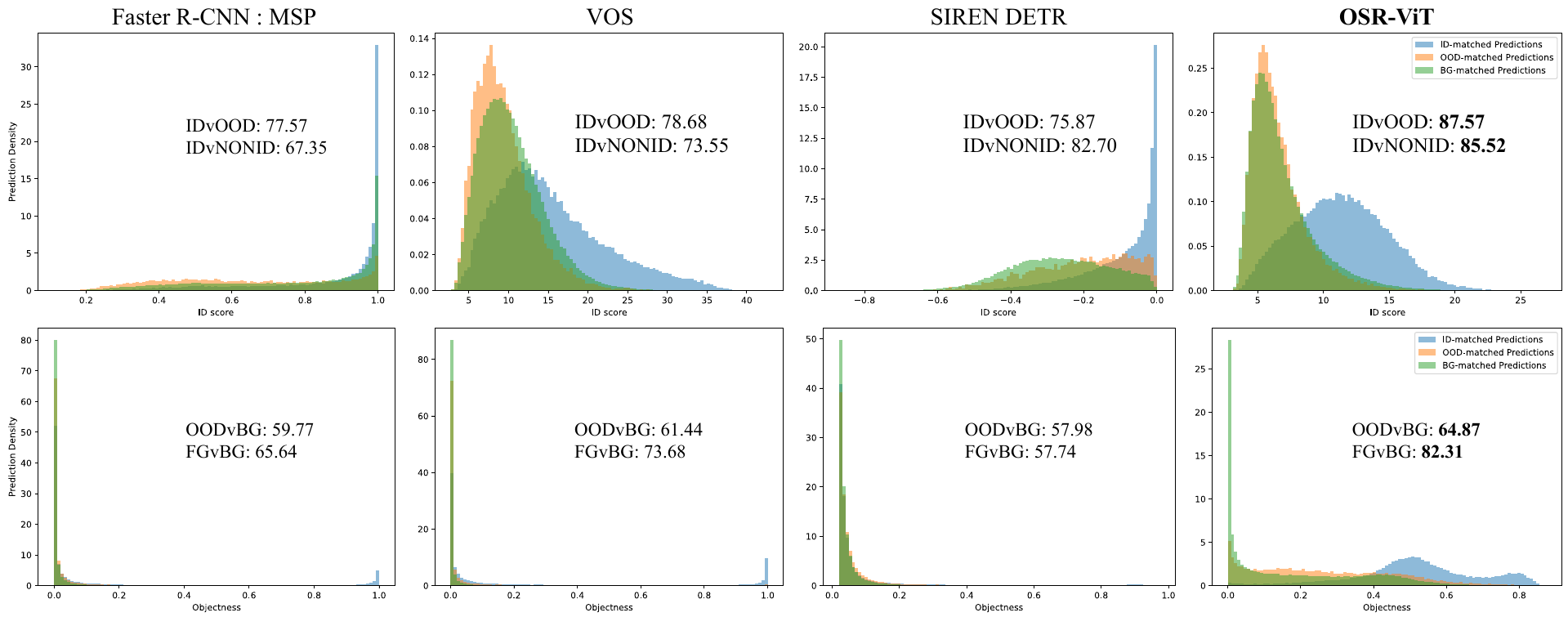}
   \vspace{-6mm}
   \caption{ID-score (top row) and objectness (bottom row) histograms of predictions matched to ID/OOD/BG ground truths from the VOC$\rightarrow$COCO task. We include the corresponding AUROC scores from each model to clarify the quantitative performance.}
   \label{fig:histograms}
   \vspace{-3mm}
\end{figure}
Figure \ref{fig:histograms} shows the histograms of ID scores from the ID, OOD, and BG-matched test-set proposals from four different models trained on the VOC$\rightarrow$COCO task. This histograms provide a visualization of what the AUROC metrics indicate. The top row shows the distributions of ID scores. In this case, we want the ID-matched predictions (the blue distribution) to have the largest scores (i.e., be the furthest right). Note that our OSR-ViT method achieves the best ID separability and shows better calibration compared to MSP \cite{MSP} and SIREN \cite{SIREN}. The bottom row shows the distributions of objectness. Here, we want the objectness of ID \textit{and} OOD-matched predictions to be higher than the background-matched predictions. Note that the baseline methods struggle to separate OOD objects and background. Our method, however, with its standalone THPN proposal network \cite{THPN} learns a much better objectness representation allowing for superior OOD vs. BG and FG vs. BG separability.

\subsection*{C. Discussion of Parameter Count}
\begin{table}[t]
\caption{Model parameter count analysis on VOC$\rightarrow$COCO task.}
\vspace{-3mm}
\centering
\resizebox{0.75\linewidth}{!}{

\begin{tabular}{l|lccc}
\toprule
Training                                            & Model                  & Params (M) & ID-mAP        & AOSP        \\ \hline\hline &&&\tabularnewline[-2.2ex]
\multirow{5}{*}{\shortstack[l]{Plain\\ Supervised}} & Faster R-CNN (RN50)    & 41.2       & 31.1          & 18.0          \\
                                                    & Faster R-CNN (RN152)   & 75.9       & 34.2          & 17.1          \\
                                                    & OLN (RN50)             & 41.2       & 30.0          & 18.8          \\
                                                    & OLN (RN152)            & 75.9       & 33.9          & 17.3          \\
                                                    & Deformable DETR (RN50) & 39.8       & \textbf{34.6} & 10.1          \\ \hline &&&\tabularnewline[-2.2ex]
VOS                                                 & Faster R-CNN (RN50)    & 41.2       & 31.5          & 18.6          \\ \hline &&&\tabularnewline[-2.2ex]
\multirow{2}{*}{SIREN}                              & Faster R-CNN (RN50)    & 41.2       & 31.3          & 17.3          \\
                                                    & Deformable DETR (RN50) & 39.8       & 33.6          & 12.0          \\ \hline &&&\tabularnewline[-2.2ex]
ORE                                                 & Faster R-CNN (RN50)    & 41.2       & 28.0          & 18.3          \\ \hline &&&\tabularnewline[-2.2ex]
OW-DETR                                             & Deformable DETR (RN50) & 39.8       & 30.2          & 10.7          \\ \hline &&&\tabularnewline[-2.2ex]
PROB                                                & Deformable DETR (RN50) & 39.8       & 32.5          & 12.6          \\ \hline &&&\tabularnewline[-2.2ex]
\multirow{3}{*}{OSR-ViT}                               & THPN(RN50)+DINOv2-S    & 63.3       & 30.2          & 23.6          \\
                                                    & THPN(RN50)+DINOv2-B    & 128.3      & 31.4          & 25.0          \\
                                                    & THPN(RN50)+DINOv2-L    & 346.6      & 31.5          & \textbf{25.1} \\
\bottomrule
\end{tabular}

}
\label{tab:param_count}
\end{table}
One downside of using foundational ViT models in a detector is that they are fairly large compared to many CNN-based alternatives. Thus, a fair question is: is OSR-ViT's superior performance is simply due to its larger parameter count? In this section, we investigate this by training some of the baselines with the much larger ResNet-152 \cite{ResNet} backbone (compared to the standard ResNet-50 that is commonly used in the literature). Results are contained in Table \ref{tab:param_count}. We find that simply increasing the number of parameters of a CNN-based fully-supervised method improves ID-mAP, but does \textit{NOT} improve AOSP (open-set performance). In fact, in our tests, the larger CNNs lead to diminished AOSP. We posit that adding parameters to an inherently ID-biased model will only make the model more ID-biased. Even our lightest weight OSR-ViT with a DINOv2-S achieves far better OSODD performance than the upscaled baselines. It should be noted that we also tried to train different baselines with the larger RN152 backbone but ran into issues. For example, DETR and VOS training was too slow, and the SIREN model did not converge to a good solution. This further supports our claim that the OSR-ViT's modular off-the-shelf-capable framework is much less of a headache to optimize because the foundational models require a minimal finetuning stage to be effective. 

\subsection*{D. Impact of THPN's $\lambda_{CLS}$ Parameter}
\begin{table}[t]
\caption{Impact of the THPN $\lambda_{CLS}$ parameter.}
\vspace{-3mm}
\centering
\resizebox{0.45\linewidth}{!}{

\begin{tabular}{l@{\hskip 2.2mm}l@{\hskip 2.2mm}ccc}
\toprule
Data                                                         & $\lambda_{CLS}$ & AOSP          & ID-mAP        & CA-AR         \\ \hline\hline &&&\tabularnewline[-2.2ex]
\multirow{4}{*}{\shortstack[l]{VOC $\rightarrow$\\ COCO}}    & 0                    & 23.8          & 29.1          & \textbf{43.7} \\
                                                             & 0.10                 & \textbf{25.0} & 31.4          & 43.2          \\
                                                             & 0.25                 & 23.9          & 32.4          & 42.7          \\
                                                             & 0.50                 & 21.5          & \textbf{32.9} & 41.1          \\ \hline
\multirow{4}{*}{\shortstack[l]{VOC75 $\rightarrow$\\ COCO}}  & 0                    & \textbf{23.4} & 28.3          & \textbf{42.6} \\
                                                             & 0.10                 & 23.1          & 30.8          & 42.2          \\
                                                             & 0.25                 & 21.3          & 31.5          & 40.9          \\
                                                             & 0.50                 & 18.9          & \textbf{31.6} & 39.2          \\ \hline
\multirow{4}{*}{\shortstack[l]{VOC50 $\rightarrow$\\ COCO}}  & 0                    & \textbf{23.1} & 27.5          & \textbf{41.4} \\
                                                             & 0.10                 & 22.8          & 29.6          & 40.7          \\
                                                             & 0.25                 & 20.8          & \textbf{30.2} & 39.5          \\
                                                             & 0.50                 & 18.1          & 30.0          & 37.7          \\ \hline
\multirow{4}{*}{\shortstack[l]{VOC25 $\rightarrow$\\ COCO}}  & 0                    & 20.1          & 24.5          & \textbf{38.5} \\
                                                             & 0.10                 & \textbf{20.9} & 26.4          & 38.4          \\
                                                             & 0.25                 & 19.1          & \textbf{26.7} & 37.0          \\
                                                             & 0.50                 & 16.4          & 25.8          & 35.2          \\ \hline
\multirow{4}{*}{\shortstack[l]{COCO $\rightarrow$\\ Obj365}} & 0                    & 16.4          & 19.7          & 49.7          \\
                                                             & 0.10                 & 18.9          & 22.2          & \textbf{51.4} \\
                                                             & 0.25                 & \textbf{20.5} & 24.4          & 51.3          \\
                                                             & 0.50                 & 19.7          & \textbf{25.1} & 49.7          \\ \hline
\multirow{4}{*}{Ships}                                       & 0                    & 54.5          & 57.7          & 63.7          \\
                                                             & 0.10                 & \textbf{55.4} & 58.9          & \textbf{64.3} \\
                                                             & 0.25                 & 54.4          & 59.7          & 63.5          \\
                                                             & 0.50                 & 50.1          & \textbf{59.8} & 61.5        \\
\bottomrule
\end{tabular}

}
\label{tab:thpn_lambda_sweeps}
\end{table}
One of the noted limitations of OSR-ViT from Sec. \ref{sec:experiments} is the inferior closed-set ID-mAP compared to some of the baselines. According to our analysis, this gap is due to the natural tradeoff that must be made with an open-set proposal network. Intuitively, the less ID-biased the proposal network is, the lower the ID recall will be (but with much improved OOD recall). Because of THPN's adaptable design, we can easily adjust its bias using its $\lambda_{CLS} \in [0,1]$ parameter. Therefore, if an application requires it we can easily close the ID-mAP gap while still maintaining good overall open-set performance. Table \ref{tab:thpn_lambda_sweeps} contains the results of sweeping $\lambda_{CLS}$ across each benchmark task. As expected, the higher we set $\lambda_{CLS}$, the more likely the resulting model is to predict more ID objects, thus the higher the ID-mAP. However, we find that in some tasks, setting $\lambda_{CLS}$ too high will yield slightly worse ID-mAP. Conversely, when we set $\lambda_{CLS} \approx 0$ we achieve much better OOD proposal recall, meaning in most cases better AOSP and CA-AR.

\subsection*{E. Full Limited Data Benchmark Results}
\begin{table*}[t]
\caption{Results on semi-supervised VOC$\rightarrow$COCO tasks.}
\vspace{-3mm}
\centering
\resizebox{0.95\linewidth}{!}{

\begin{tabular}{l|l|l|lccccccc}
\toprule
                                                              &                                                      &                                  &              &               &               &               & IDvOOD         & IDvNONID       & OODvBG         & FGvBG          \\
Data                                                          & Training                                             & Model                            & OOD Algo.    & AOSP        & ID-mAP        & CA-AR         & AUROC          & AUROC          & AUROC          & AUROC          \\ \hline\hline &&&\tabularnewline[-2.2ex]
\multirow{15}{*}{\shortstack[l]{VOC75 $\rightarrow$\\ COCO}} & \multirow{6}{*}{\shortstack[l]{Plain\\ Supervised}} & \multirow{2}{*}{Faster R-CNN}    
& Energy       & 16.3          & 28.7          & 36.3          & 73.59          & 63.63          & 59.81          & 65.26          \\
                                                              &                                                      &                                  & Mahalanobis  & 16.2          & 28.7          & 36.3          & 57.48          & 70.54          & 59.81          & 65.26          \\ \cline{3-11} &&&\tabularnewline[-2.2ex]
                                                              &                                                      & \multirow{2}{*}{OLN}         
& Energy       & 17.6          & 27.5          & 37.4          & 72.31          & 65.26          & 59.07          & 65.89          \\
                                                              &                                                      &                                  & Mahalanobis  & 17.2          & 27.5          & 37.4          & 52.90          & 67.24          & 59.07          & 65.89          \\ \cline{3-11} &&&\tabularnewline[-2.2ex]
                                                              &                                                      & \multirow{2}{*}{Deformable DETR} 
& Energy       & 8.9           & 30.4          & 31.9          & 59.48          & 70.50          & 59.74          & 58.14          \\
                                                              &                                                      &                                  & Mahalanobis  & 7.8           & 30.4          & 31.9          & 61.10          & 70.36          & 59.74          & 58.14          \\ \cline{2-11} &&&\tabularnewline[-2.2ex]
                                                              & VOS                                                  & Faster R-CNN                     & Energy       & 16.9          & 28.5          & 35.5          & 78.91          & 72.57          & 60.18          & 71.95          \\ \cline{2-11} &&&\tabularnewline[-2.2ex]
                                                              & \multirow{2}{*}{SIREN}                               & Faster R-CNN                     & SIREN-KNN    & 16.4          & 28.8          & 35.7          & 82.93          & 77.52          & 58.24          & 63.67          \\
                                                              &                                                      & Deformable DETR                  & SIREN-KNN    & 8.8           & 29.4          & 31.4          & 78.92          & 83.54          & 57.70          & 57.25          \\ \cline{2-11} &&&\tabularnewline[-2.2ex]
                                                              & ORE                                                  & Faster R-CNN                     & Energy       & 17.5          & 26.1          & 34.7          & 76.80          & 76.07          & 54.17          & 62.97          \\ \cline{2-11} &&&\tabularnewline[-2.2ex]
                                                              & OW-DETR                                              & Deformable DETR                  & Direct Pred. & 8.9           & 25.3          & 28.5          & -              & -              & -              & -              \\ \cline{2-11} &&&\tabularnewline[-2.2ex]
                                                              & PROB                                                 & Deformable DETR                  & Direct Pred. & 11.2          & 27.2          & 30.2          & -              & -              & -              & -              \\ \cline{2-11} &&&\tabularnewline[-2.2ex]
                                                              & \multirow{3}{*}{OSR-ViT}                                & THPN+DINOv2-S                    & Energy       & 22.4          & 30.2          & \textbf{42.2} & 85.56          & 84.09          & 63.43          & 81.10          \\
                                                              &                                                      & THPN+DINOv2-B                    & Energy       & \textbf{23.1} & \textbf{30.8} & \textbf{42.2} & \textbf{87.74} & \textbf{84.62} & 62.31          & 81.05          \\
                                                              &                                                      & THPN+DINOv2-L                    & Energy       & 23.0          & 30.7          & \textbf{42.2} & 85.87          & 83.18          & \textbf{65.43} & \textbf{82.56} \\ \hline\hline &&&\tabularnewline[-2.2ex]
\multirow{15}{*}{\shortstack[l]{VOC50 $\rightarrow$\\ COCO}} & \multirow{6}{*}{\shortstack[l]{Plain\\ Supervised}} & \multirow{2}{*}{Faster R-CNN}    
& Energy       & 15.0          & 25.6          & 34.9          & 74.80          & 62.60          & 60.37          & 65.45          \\
                                                              &                                                      &                                  & Mahalanobis  & 15.1          & 25.6          & 34.9          & 60.41          & 73.47          & 60.37          & 65.45          \\ \cline{3-11} &&&\tabularnewline[-2.2ex]
                                                              &                                                      & \multirow{2}{*}{OLN}             
& Energy       & 16.0          & 24.8          & 36.0          & 72.05          & 62.44          & 60.35          & 66.36          \\
                                                              &                                                      &                                  & Mahalanobis  & 15.7          & 24.8          & 36.0          & 53.12          & 69.07          & 60.35          & 66.36          \\ \cline{3-11} &&&\tabularnewline[-2.2ex]
                                                              &                                                      & \multirow{2}{*}{Deformable DETR} 
& Energy       & 9.2           & 22.7          & 29.6          & 57.10          & 66.96          & 59.46          & 59.97          \\
                                                              &                                                      &                                  & Mahalanobis  & 8.2           & 22.7          & 29.6          & 64.52          & 73.10          & 59.46          & 59.97          \\ \cline{2-11} &&&\tabularnewline[-2.2ex]
                                                              & VOS                                                  & Faster R-CNN                     & Energy       & 15.1          & 25.2          & 33.6          & 78.40          & 68.86          & 60.04          & 71.13          \\ \cline{2-11} &&&\tabularnewline[-2.2ex]
                                                              & \multirow{2}{*}{SIREN}                               & Faster R-CNN                     & SIREN-KNN    & 15.1          & 25.6          & 34.0          & 83.17          & 76.42          & 59.85          & 64.30          \\
                                                              &                                                      & Deformable DETR                  & SIREN-KNN    & 8.1           & 24.2          & 28.7          & 78.42          & 84.75          & 60.29          & 60.31          \\ \cline{2-11} &&&\tabularnewline[-2.2ex]
                                                              & ORE                                                  & Faster R-CNN                     & Energy       & 14.4          & 22.9          & 32.9          & 76.58          & 75.75          & 55.09          & 64.14          \\ \cline{2-11} &&&\tabularnewline[-2.2ex]
                                                              & OW-DETR                                              & Deformable DETR                  & Direct Pred. & 8.9           & 21.9          & 25.8          & -              & -              & -              & -              \\ \cline{2-11} &&&\tabularnewline[-2.2ex]
                                                              & PROB                                                 & Deformable DETR                  & Direct Pred. & 9.3           & 22.9          & 28.4          & -              & -              & -              & -              \\ \cline{2-11} &&&\tabularnewline[-2.2ex]
                                                              & \multirow{3}{*}{OSR-ViT}                                & THPN+DINOv2-S                    & Energy       & 22.0          & 28.8          & \textbf{40.7} & 84.56          & 83.31          & 62.65          & 80.57          \\
                                                              &                                                      & THPN+DINOv2-B                    & Energy       & \textbf{22.8} & \textbf{29.6} & \textbf{40.7} & \textbf{85.94} & \textbf{84.57} & 65.31          & 81.99          \\
                                                              &                                                      & THPN+DINOv2-L                    & Energy       & 22.2          & 28.9          & \textbf{40.7} & 85.36          & 83.99          & \textbf{66.43} & \textbf{82.77} \\ \hline\hline &&&\tabularnewline[-2.2ex]
\multirow{15}{*}{\shortstack[l]{VOC25 $\rightarrow$\\ COCO}} & \multirow{6}{*}{\shortstack[l]{Plain\\ Supervised}} & \multirow{2}{*}{Faster R-CNN}    
& Energy       & 12.5          & 20.5          & 32.3          & 70.53          & 60.49          & 60.92          & 65.39          \\
                                                              &                                                      &                                  & Mahalanobis  & 12.7          & 20.5          & 32.3          & 63.09          & 72.83          & 60.92          & 65.39          \\ \cline{3-11} &&&\tabularnewline[-2.2ex]
                                                              &                                                      & \multirow{2}{*}{OLN}            
& Energy       & 13.1          & 20.2          & 33.6          & 72.65          & 62.73          & 60.64          & 66.27          \\
                                                              &                                                      &                                  & Mahalanobis  & 12.7          & 20.2          & 33.6          & 52.30          & 67.40          & 60.64          & 66.27          \\ \cline{3-11} &&&\tabularnewline[-2.2ex]
                                                              &                                                      & \multirow{2}{*}{Deformable DETR} 
& Energy       & 8.2           & 15.8          & 26.7          & 54.50          & 65.07          & 59.46          & 59.74          \\
                                                              &                                                      &                                  & Mahalanobis  & 7.3           & 15.8          & 26.7          & 64.52          & 74.86          & 59.46          & 59.74          \\ \cline{2-11} &&&\tabularnewline[-2.2ex]
                                                              & VOS                                                  & Faster R-CNN                     & Energy       & 12.6          & 20.2          & 31.4          & 78.26          & 65.72          & 60.21          & 68.63          \\ \cline{2-11} &&&\tabularnewline[-2.2ex]
                                                              & \multirow{2}{*}{SIREN}                               & Faster R-CNN                     & SIREN-KNN    & 13.2          & 20.5          & 32.4          & 81.58          & 74.45          & 60.45          & 64.56          \\
                                                              &                                                      & Deformable DETR                  & SIREN-KNN    & 7.8           & 14.8          & 26.3          & 77.46          & 79.03          & 58.86          & 59.96          \\ \cline{2-11} &&&\tabularnewline[-2.2ex]
                                                              & ORE                                                  & Faster R-CNN                     & Energy       & 12.7          & 18.9          & 30.5          & 77.73          & 76.88          & 55.07          & 64.10          \\ \cline{2-11} &&&\tabularnewline[-2.2ex]
                                                              & OW-DETR                                              & Deformable DETR                  & Direct Pred. & 5.6           & 14.7          & 22.3          & -              & -              & -              & -              \\ \cline{2-11} &&&\tabularnewline[-2.2ex]
                                                              & PROB                                                 & Deformable DETR                  & Direct Pred. & 6.3           & 15.7          & 24.8          & -              & -              & -              & -              \\ \cline{2-11} &&&\tabularnewline[-2.2ex]
                                                              & \multirow{3}{*}{OSR-ViT}                                & THPN+DINOv2-S                    & Energy       & 20.6          & 25.7          & \textbf{38.4} & 84.86          & 82.10          & 57.42          & 78.34          \\
                                                              &                                                      & THPN+DINOv2-B                    & Energy       & \textbf{20.9} & \textbf{26.4} & \textbf{38.4} & \textbf{86.97} & \textbf{83.88} & \textbf{61.29} & \textbf{80.54} \\
                                                              &                                                      & THPN+DINOv2-L                    & Energy       & 20.8          & 25.8          & \textbf{38.4} & 86.46          & 81.80          & 59.11          & 79.97 \\
\bottomrule
\end{tabular}

}
\label{tab:semi_supervised}
\end{table*}
In this section, we provide the full tabular results (see Table \ref{tab:semi_supervised}) for the Limited Data Benchmark detailed in Sec. \ref{subsec:limited_data_benchmark}. The key takeaway from this experiment is that our OSR-ViT framework maintains performance in low-data settings far better than a fully-supervised model can. Specifically, note that an OSR-ViT trained on 25\% of the VOC training annotations outperforms all baselines trained on 100\% of the annotations!

\subsection*{F. Prediction Samples}
\begin{figure}[h]
  \centering
   \includegraphics[width=1.0\linewidth]{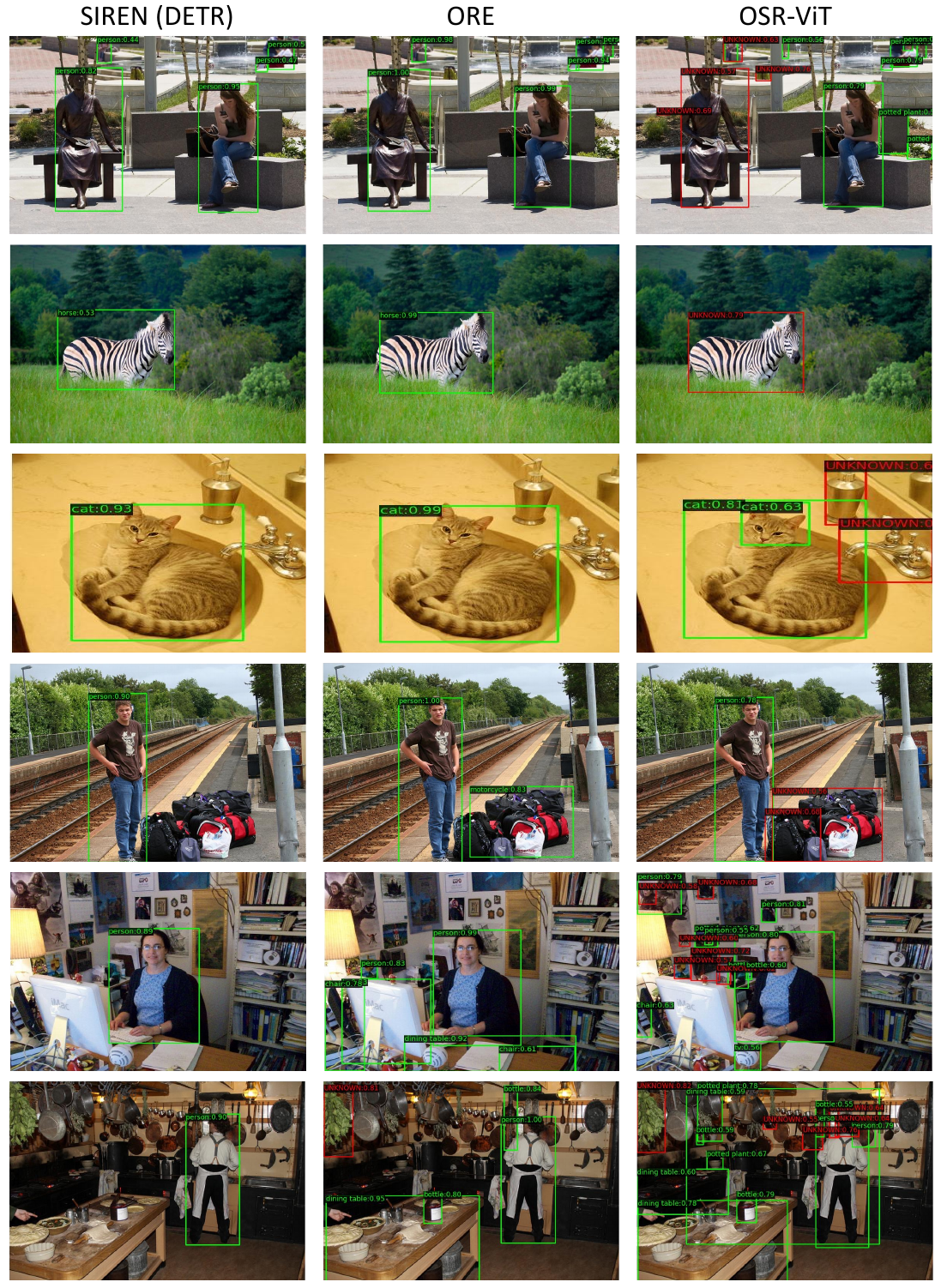}
   \vspace{-5mm}
   \caption{Predicted samples on the VOC$\rightarrow$COCO task. Note how our OSR-ViT model detects more OOD objects and achieves better fine-grained OOD separability on cases like zebra vs. horse and statue vs. person. }
   \label{fig:samples_voccoco}
\end{figure}
\begin{figure}[h]
  \centering
   \includegraphics[width=1.0\linewidth]{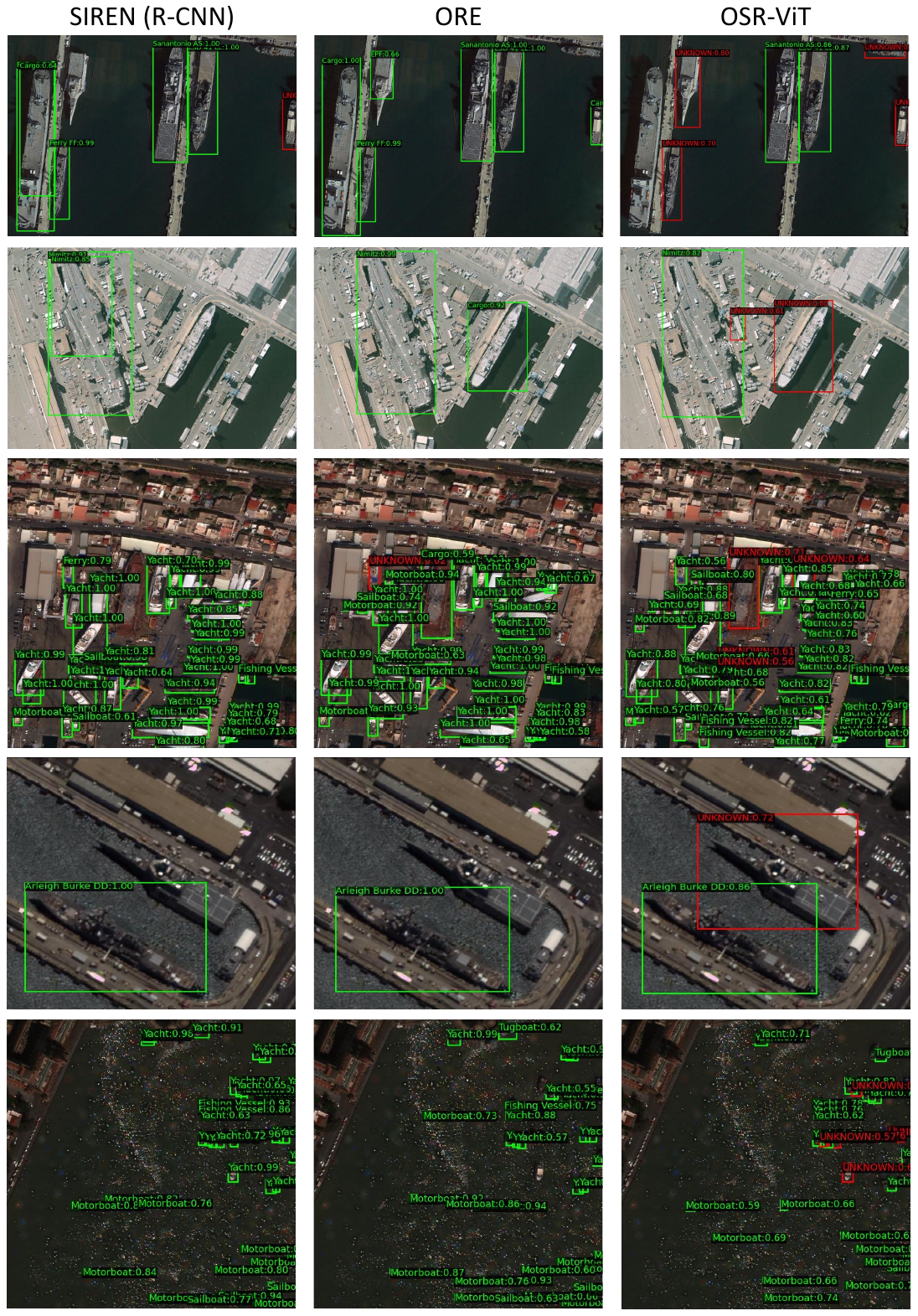}
   \vspace{-5mm}
   \caption{Predicted samples on the Ships task. Note how our OSR-ViT model can generalize to OOD ship classes and does a better job at separating ID vs. OOD ships.}
   \label{fig:samples_ships}
\end{figure}
Fig. \ref{fig:samples_voccoco} and Fig. \ref{fig:samples_ships} contain prediction samples for the VOC$\rightarrow$COCO and Ships tasks, respectively. OSR-ViT shows better OOD recall and superior ID vs. OOD separability characteristics. Note that we use a small validation set to determine the ``optimal'' deployment thresholds for making these figures.

\subsection*{G. Task Details}
\begin{table}[h]
\caption{Comparison of benchmark tasks.}
\vspace{-3mm}
\centering
\resizebox{0.75\linewidth}{!}{

\begin{tabular}{lc|cc|ccc}
\toprule
                            & \multicolumn{1}{l|}{} & \multicolumn{2}{c|}{Train} & \multicolumn{3}{c}{Test}    \\
Benchmark                   & ID Classes            & Images     & Instances     & Images & Instances & OOD \\ \hline\hline &&&\tabularnewline[-2.2ex]
VOC$\rightarrow$COCO        & 20                    & 16551      & 47223         & 5000   & 36781     & 42\%     \\
VOC75$\rightarrow$COCO      & 20                    & 14593      & 35419         & 5000   & 36781     & 42\%     \\
VOC50$\rightarrow$COCO      & 20                    & 11879      & 23610         & 5000   & 36781     & 42\%     \\
VOC25$\rightarrow$COCO      & 20                    & 7718       & 11804         & 5000   & 36781     & 42\%     \\
COCO$\rightarrow$Objects365 & 80                    & 118287     & 860001        & 40000  & 622194    & 61\%     \\
Ships                       & 41                    & 2098       & 1985          & 550    & 2949      & 23\% \\
\bottomrule
\end{tabular}

}
\label{tab:benchmarks}
\end{table}

Table \ref{tab:benchmarks} provides details about the different benchmark tasks that we consider in this paper. Note that we study a far more diverse range of tasks than other literature in the field. 


\end{document}